\documentclass{article}
\usepackage{iclr2027_conference,times}

\usepackage[T1]{fontenc}
\usepackage[utf8]{inputenc}

\usepackage{amsmath}    % align, equation, \text, \mathbb ...
\usepackage{amssymb}    % \mathbb, \mathcal, \nleq ...
\usepackage{amsthm}     % theorem environments and \begin{proof}
\usepackage{bm}         % \bm for bold math symbols
\usepackage{mathtools}  % \DeclarePairedDelimiter, better \colon etc.
\usepackage{graphicx}
\usepackage{algorithm}
\usepackage{algpseudocode}
\usepackage{enumitem}
\usepackage{booktabs,longtable,graphicx}
\usepackage[colorlinks=true,linkcolor=blue,citecolor=blue,urlcolor=blue]{hyperref}
\usepackage[capitalise,noabbrev]{cleveref}
\usepackage{wrapfig}
\theoremstyle{plain}
\newtheorem{theorem}{Theorem}[section]
\newtheorem{lemma}[theorem]{Lemma}
\newtheorem{proposition}[theorem]{Proposition}

\theoremstyle{definition}

\theoremstyle{remark}

\newcommand{\bx}{\bm{x}}

\title{Graph-Localized Coverage for Conformal Prediction on Inductive Graphs}

\author{
Jinwen Xu \\
\textit{School of ECE} \\
\textit{University of Georgia} \\
Athens, GA 30602, USA \\
\texttt{jinwen.xu@uga.edu}
\And
Qin Lu \\
\textit{School of ECE} \\
\textit{University of Georgia} \\
Athens, GA 30602, USA \\
\texttt{qin.lu@uga.edu}
\And
Georgios B. Giannakis \\
\textit{Dept. of ECE} \\
\textit{University of Minnesota} \\
Minneapolis, MN 55455, USA \\
\texttt{georgios@umn.edu}
}

\iclrfinalcopy

\begin{document}
\maketitle

% Replace the ICLR camera-ready header with a preprint header.
\lhead{Preprint}

\begin{abstract}
Conformal prediction (CP) provides marginal coverage guarantees, but marginal
coverage does not determine where prediction errors occur on a graph.
Queries from different graph neighborhoods can exhibit substantially
different levels of uncertainty even when aggregate coverage remains near its target.
We introduce GLocalCP, an online CP framework for
{\it inductive} node classification and edge prediction that combines marginal
and graph-local calibration with uncertainty-adaptive prediction sets. Specifically, 
GLocalCP puts forth a novel {\it uncertainty-adaptive} nonconformality score that relies on the  mutual information obtained from Monte-Carlo (MC) dropout based prediction. Then, it augments a global online threshold with a kernel-based local correction that propagates coverage feedback among similar graph queries. 
On the theoretical horizon, we establish pathwise $O(N^{-1/2})$ marginal calibration error without
exchangeability or stationarity and, for the exact-kernel formulation,
an explicit $O(N^{-1/2})$ graph-localized calibration guarantee.
Experiments on seven node-classification and seven edge-prediction
benchmarks show that similar marginal coverage can conceal substantially
different local reliability. Our proposed GLocalCP maintains coverage near the target
while consistently reducing graph-local calibration error.
Uncertainty scaling in GLocalCP further reduces prediction-set size across all
evaluated node and edge datasets while leaving marginal coverage
essentially unchanged.
\end{abstract}

\section{Introduction}
\label{sec:introduction}

Graph neural networks (GNNs) \citep{kipf2017semi,velickovic2018graph} are increasingly deployed for prediction over
relational data, where reliable uncertainty quantification is needed alongside
accurate predictions~\citep{hamilton2017sage}. Conformal prediction (CP)
provides model-agnostic prediction sets with distribution-free marginal
coverage under exchangeability
\citep{vovk2005conformal,angelopoulos2021gentle,romano2020classification},
and recent work has extended CP to graph-structured prediction
\citep{zargarbashi2023conformal,lunde2023conformal,
zargarbashi2024conformalinductive}. Existing graph CP methods operate under
different prediction settings. Some focus on a fixed transductive graph,
while others accommodate inductive queries through specific
calibration--test constructions. Inductive prediction alone, however, does
not address distribution shift in a sequential graph stream. As new nodes
arrive, the observed graph, query contexts, and their distribution may evolve,
so a calibration rule appropriate earlier will not remain
calibrated later. Online CP addresses this temporal adaptation by updating a
{\it global} calibration threshold from sequential coverage feedback
\citep{gibbs2021aci,angelopoulos2024decaying,angelopoulos2024pid}. {These methods, however, are developed for general sequential prediction rather than inductively evolving graphs, where both the graph structure and query contexts change as new nodes arrive.}

Such a {\it global} adaptation can only yield {\it marginal} calibration that 
controls the aggregate error rate, but can not characterize how errors are distributed across
the graph-query space \citep{foygel2021limits}.
A predictor can achieve the target coverage on average while persistently
under-covering some types of node or edge queries and over-covering others.
This motivates calibration procedures that account for {\it local} context
\citep{guan2023localized}.
The distinction is particularly relevant on graphs, where nonconformity
scores and prediction-set efficiency can depend on network structure
\citep{huang2023uncertainty}.
Moreover, predictive uncertainty can vary substantially across queries,
motivating prediction sets whose sizes adapt to query difficulty
\citep{romano2020classification}.
These considerations motivate an online conformal framework that adapts
not only over time, but also to \emph{where} coverage errors occur and
\emph{how uncertain} each query is.

We introduce \textbf{GLocalCP} (Graph-Localized Conformal Prediction), an
uncertainty-adaptive online calibration framework for inductive node
classification and edge prediction. GLocalCP augments a global online
calibration state with two complementary mechanisms. First, a kernel-based
local state associates realized coverage errors with their graph-query
representations and propagates feedback among similar queries, enabling
calibration to respond to localized as well as aggregate errors. Second, Monte Carlo dropout \citep{gal2016dropout} quantifies
query-specific predictive uncertainty, from which we construct a
predictable mutual-information (MI) score
\citep{houlsby2011bayesian} scale adapts prediction-set
allocation. For uncertainty scales bounded away from zero, we establish
pathwise $O(N^{-1/2})$ (where $N$ is the number of online queries) marginal calibration without exchangeability or
stationarity. Under a fixed normalized and uniformly positive kernel, the
exact-kernel update further achieves an $O(N^{-1/2})$ uniform graph-local
calibration bound.

Experiments on seven graph benchmarks for both node classification and
held-out edge prediction support the two mechanisms. GLocalCP maintains
marginal coverage near the $90\%$ target while attaining substantially
smaller graph-local calibration gaps than graph CP baselines, including
cases in which competing methods achieve similar marginal coverage. Controlled
ablations show that the kernel-local feedback drives the reduction in local
calibration error, whereas MI scaling reduces mean prediction-set size on
all evaluated node and edge datasets while preserving near-target marginal
coverage. The local-calibration improvements further persist across
alternative GNN backbones, nonconformity scores, and edge-candidate
distributions.

The contributions of this work are summarized in the following three aspects.
\begin{itemize}
    \item 
    We formulate graph-localized calibration over inductive node and edge
    representations and introduce kernel-localized feedback that adapts
    sequential coverage errors across similar graph queries.

    \item 
    We introduce a predictable MI-based uncertainty scale that adapts
    prediction-set allocation to query difficulty while preserving a common
    target miscoverage level.

    \item 
    We establish $O(N^{-1/2})$ pathwise marginal and graph-localized calibration guarantees without exchangeability or stationarity,
    and empirically separate the roles of local feedback and uncertainty
    scaling across inductive node and edge streams.
\end{itemize}

\section{Problem Formulation and Preliminaries}
\label{sec:problem}
Consider an inductive graph setting, where $\mathcal G_n:=(\mathcal V_n,\mathcal E_n, {\bf X}_n)$ denote the observed graph information at round $n$. Here, $\mathcal V_n$ and $\mathcal E_n$ are
the available nodes and edges and
${\bf X}_n:=\{\bx_v:v\in\mathcal V_n\}$ contains the corresponding nodal features.
We consider an evolving graph
$\mathcal G_1\subseteq\mathcal G_2\subseteq\cdots$, where new nodes and
edges may arrive over time. The goal is to predict the label $y_{v}\in\mathcal Y:=\{1,\ldots,C\}$ of a new node $v \in \Delta {\cal V}_n$ ($\Delta {\cal V}_n$ collects new nodes at round $n$) or the existence $y_e\in \{0,1\}$ of an edge $e:=(v,v')$ connecting nodes $v$ and $v'$.
For both tasks, we will rely on an inductive GNN encoder ${\bf g}_\theta$ (e.g., GraphSAGE~\citep{hamilton2017sage}) that   maps
each node from its features and currently observed neighborhood to an embedding
\begin{align}
{\bf z}_{n} (v)
=
{\bf g}_\theta\!\left(\bx_v,\mathcal N_v(\mathcal G_n)\right)
\in\mathbb R^{d_z},
\qquad v\in\mathcal V_n. \label{eq:node_embedding}
\end{align}
Since ${\bf g}_\theta$ is a shared mapping rather than a collection of
node-specific embeddings, it can be applied to nodes arriving after
training \citep{chen2018fastgcn,zeng2020graphsaint}. For node classification, this embedding is then directly mapped to class probabilities via a node classifier $\boldsymbol{\rho}_{\mathrm{node}}$ as
\begin{align}
{\mathbf p}_{n}(v)
=\boldsymbol{\rho}_{\mathrm{node}}({\bf z}_{n}(v))\in\Delta^{C-1}, \label{eq:node_prob}
\end{align}
where $\Delta^{C-1}$ denotes the $C$-class probability simplex. For edge prediction, we will first obtain the edge feature based on the corresponding node embeddings via
\begin{align}
{\bf z}_{n} (e)
=
\left[
\left({\bf z}_{n}(v)+{\bf z}_{n}(v')\right)^\top,\,
\left|{\bf z}_{n}(v)-{\bf z}_{n}(v')\right|^\top,\,
\left({\bf z}_{n}(v)\odot {\bf z}_{n}(v')\right)^\top
\right]^\top \label{eq:edge_embedding}
\end{align}
where $\odot$ denotes the element-wise (Hadamard) product. The resulting
edge feature is further fed into the edge classifier to yield
\begin{align}
{\mathbf p}_{n}(e)
=\boldsymbol{\rho}_{\mathrm{edge}}({\bf z}_{n}(e))\in\Delta^1.   \label{eq:edge_prob}
\end{align}

{\bf Objectives.} We pre-train the graph encoder ${\bf g}_\theta$ as well as the node and edge classifiers $\{ \boldsymbol{\rho}_{\mathrm{node}}, \boldsymbol{\rho}_{\mathrm{edge}}\}$ on the initial graph ${\cal G}_0$ along with node labels ${\bf y}_0:= \{y_v: v\in {\cal V}_0\}$ . For uncertainty awareness, we will go beyond the {\it point} prediction to obtain a {\it prediction set} $\mathcal C_n (v)$ for the label of node $v$ or $\mathcal C_n (e)$ for the existence of an edge $e:=(v,v')$ connecting nodes $v$ and $v'$. Here, the set $\mathcal C_n (\cdot)$ is sought before observing $y_n\in\mathcal Y$. Because the two tasks differ only in the feature representation and
the classifier, we hereafter write ${\bf z}_n$ and ${\bf p}_n$ in place of
${\bf z}_n(v)$/${\bf z}_n(e)$ and ${\bf p}_n(v)$/${\bf p}_n(e)$ for brevity. Also, since the prediction sets here are constructed based on latent embeddings, we will use $\mathcal C_n ({\bf z})$ in place of $\mathcal C_n (v)$/$\mathcal C_n (e)$ in the ensuing discussions.

\subsection{Conformal prediction (CP)}
CP is a prevailing framewrok that constructs prediction sets with coverage guarantees~\citep{vovk2005conformal}. Given a target miscoverage level $\alpha\in(0,1)$, CP relies on a {\it nonconformity} score $s({\bf z}, y)$, which measures how a candidate label $y\in\mathcal Y$ agrees with the prediction of ${\bf z}$ given by a pretrained model. Specifically,
smaller values indicate greater agreement with the model prediction.
Given scores $\{s({\bf z}_i,y_i)\}_{i=1}^{N_{\mathrm{cal}}}$ evaluated at held out calibration data, split CP
chooses a threshold $q$ as the corresponding empirical $(1-\alpha)$-quantile
and obtain the prediction set
\begin{align}
\mathcal C_n({\bf z})
:=
\{y\in\mathcal Y:s({\bf z},y)\le q\}.
\end{align}
Under exchangeability of the calibration and test data, split CP
provides the following finite-sample marginal coverage guarantee
\begin{align}
\mathbb P\!\left\{y_n\in\mathcal C_n({\bf z})\right\}\ge 1-\alpha.
\end{align}

We consider two standard nonconformity scores for classification tasks. The first one is the Least Ambiguous Classifier (LAC), which uses $s({\bf z},y)=1-p_{n,y}$, where $p_{n,y}$ is the predicted probability of class $y$ for a node or an edge. Alternatively, {the adaptive prediction sets (APS) score orders labels by decreasing predictive probability. With $\pi_n(j)$ denoting the label with the $j$th largest predicted probability and $\operatorname{rank}_n(y)$ denoting the position of label $y$ in this ordering, the APS score is given by $s({\bf z},y)=\sum_{j=1}^{\operatorname{rank}_n(y)}p_{n,\pi_n(j)}$.} Both scores lie in $[0,1]$.

\paragraph{Existing works on graph-based and online CP.}
Existing graph-based CP methods exploit graph structure under different
prediction settings. For node classification, DAPS~\citep{zargarbashi2023conformal}
and SNAPS~\citep{song2024snaps} are primarily developed in the transductive
setting, using graph neighborhoods and feature similarity to refine
nonconformity scores. NAPS-U~\citep{clarkson2023naps}, in contrast, supports
inductive node prediction by reweighting calibration nodes according to
their proximity to a newly observed test node. For link prediction,
CLP-CQR and CLP-SCQR~\citep{zhao2024clp} conformalize link predictions under
an inductive calibration--test construction, while RR-CQR and
RR-Mondrian~\citep{zhang2025rr} incorporate graph-dependent residual
reweighting or partitioning into calibration. These inductive methods rely
on specific relationships between calibration and test queries, which may
change as the graph and query distribution evolve over time. A related line
of work combines online CP with scalable Gaussian processes (GPs):
\citet{xu2025icassp} use sequential coverage feedback to adapt conformal
thresholds for online GP prediction, while \citet{xu2025graph}
extend this approach to graph-structured data using an ensemble of
graph-aware GPs. Although these methods adapt to sequentially arriving data,
they do not specifically address inductive node and edge queries in an
evolving graph. In our setting, the true label is revealed after each
prediction, allowing us to use sequential coverage feedback to adapt
prediction sets for subsequent graph queries.

Aggregate coverage, however, does not characterize how coverage errors are
distributed across the query space. A predictor can achieve the target
coverage on average while exhibiting persistent under-coverage or
over-coverage for particular groups of node or edge queries. This distinction
is especially important on graphs, where queries with similar structural and
predictive contexts can occupy localized regions of the learned representation
space. We therefore go beyond aggregate coverage and study
\emph{graph-localized calibration}, which asks whether coverage remains close
to the target within such local regions. We formalize this notion and develop
its online control in Section~\ref{sec:local}.

\section{GLocalCP: Uncertainty-Adaptive Online Calibration}
\label{sec:method}

%The previous section formulates marginal and graph-localized calibration for inductive graph streams. 
We now develop GLocalCP by exploiting two
complementary signals available before each label is revealed. Stochastic
GNN predictions characterize the uncertainty of the current node or edge
query, while the inductive representation $z_n$ characterizes its graph
context and hence its similarity to previously observed queries. GLocalCP
uses the former to adapt the nonconformality score and the latter to localize
online calibration feedback.

\subsection{Uncertainty-Adaptive nonconformality Score}
\label{sec:scaling}

% %Consider first a node query $x_n=v_n$. Given the observed graph
% $\mathcal G_n$, the inductive GNN encodes $v_n$ from its features and
% currently observed neighborhood and produces a predictive distribution over
% $\mathcal Y=\{1,\ldots,C\}$. For an edge query $x_n=(u_n,v_n)$, the same
% encoder produces endpoint representations, whose symmetric pair
% representation defined in Section~\ref{sec:problem} is passed to the edge
% classifier to obtain a binary predictive distribution over
% $\mathcal Y=\{0,1\}$. Thus, the two tasks differ in their graph-query
% representations and prediction heads, but both provide class probabilities
% from which conformal scores are constructed.

To characterize query-specific predictive uncertainty, we apply Monte Carlo (MC)
dropout \citep{gal2016dropout} to the corresponding graph predictor. Specifically, let $\widetilde{\bf p}_n^{(m)}$ represent the class probabilities from the $m$th dropout realization. Here,
$\widetilde{\bf p}_n^{(m)}$ is a $C\times 1$ vector for node classification and a $2\times 1$ vector for edge prediction. Then, the aggregated prediction is
\begin{align}
   \widetilde {\bf p}_{n}=M^{-1}\sum_{m=1}^{M}\widetilde{\bf p}_n ^{(m)} \label{eq:av}
\end{align}

The stochastic predictions additionally reveal uncertainty that is not
captured by their mean. We quantify their disagreement using predictive
mutual information (MI)
\begin{equation}
\operatorname{MI}_n
=
H(\widetilde{\mathbf p}_n)
-
\frac{1}{M}
\sum_{m=1}^{M}
H(\widetilde{\bf p}_n^{(m)}),
\label{eq:predictive_mi}
\end{equation}
 {where
$H(\mathbf p):=-\sum_{c\in\mathcal Y}p_c\log p_c$
denotes predictive entropy. We then define the uncertainty scale}
\begin{equation}
r_n
=
\max\left\{
\gamma_r,\,
\sqrt{
\frac{\operatorname{MI}_n}
{\log|\mathcal Y|}
}
\right\},
\qquad
r_n\in[\gamma_r,1],
\label{eq:mi_scale}
\end{equation}
where $\gamma_r>0$ bounds the scale away from zero. The normalization by
$\log|\mathcal Y|$ places the MI signal on a common scale for the
multiclass node and binary edge tasks. %Since all stochastic predictions are computed before observing $y_n$, $r_n$ is predictable.

We incorporate uncertainty directly through the scaled nonconformity score
$\bar{s}({\bf z},y):=s({\bf z},y)/r_n$. Since
$s({\bf z},y)\in[0,1]$ and $r_n\ge\gamma_r$, $\bar{s}({\bf z},y)$ is
bounded by $1/\gamma_r$. The role of $r_n$ is to adapt prediction-set
allocation to the uncertainty of each node or edge query, while leaving the
common target miscoverage level $\alpha$ unchanged. Note that MC dropout is only used to calculate the uncertainty scale $r_n$.  ${s}({\bf z},y)$ is evaluated using single-pass deterministic prediction in the previous section.

\subsection{Global and Graph-localized Feedback}
\label{sec:glocal_update}

Building on the novel uncertainty-adaptive score function, we are ready to obtain the prediction set. Given the inductive setting where the label is revealed after constructing the prediction set, we will adapt the online CP framework~\cite{gibbs2021aci} which adjusts the threshold $q_n$ based on the feedback whether the true label is covered by the prediction set.  In the conventional online CP, however, $q_n$ is a {\it global} threshold which does not account for local information over graphs. 

To allow for localized calibration, we additionally maintain $g_n ({\bf z})\in\mathcal H_k$, where $\mathcal H_k$ is a reproducing kernel Hilber space (RKHS) associated
with a positive-definite kernel function $k$. Here, we use the node/edge embeddings (cf. (\ref{eq:node_embedding}) and (\ref{eq:edge_embedding})) as the input to $g_n(\cdot)$. It is intuitive that nodes/edges with similar embeddings share the calibration threshold. Taking into account both global and local thresholds, GLocalCP constructs the conformal prediction set for a new query with embedding ${\bf z}_n$
\begin{equation}
\mathcal C_n ({\bf z}_n)
=
\left\{
y\in\mathcal Y:
\frac{s({\bf z}_n,y)}{r_n}
\le
q_n+g_n({\bf z}_n)
\right\}.
\label{eq:glocal_set}
\end{equation}

After issuing $\mathcal C_n$, the true node class or edge label $y_n$ is
revealed, yielding the centered miscoverage feedback
\begin{align}
  \epsilon_n
=
\mathbb{I}\{y_n\notin\mathcal C_n ({\bf z}_n)\}-\alpha. \label{eq:coverage_error}
\end{align}
where $\mathbb{I}(\cdot)$ is an indicator function which takes the value of $1/0$ when the statement inside is True/False. Based on this feedback, 
GLocalCP then updates the global and local thresholds as
\begin{equation}
q_{n+1}
=
q_n+\eta \epsilon_n,
\qquad
g_{n+1}
=
g_n+\eta \epsilon_n k(\cdot,{\bf z}_n),
\label{eq:glocal_update}
\end{equation}
where $\eta>0$ is the step size. The update of the global $q_n$, same as in the previous online CP framewroks, aggregates coverage
error over the stream. As for the local threshold $g_n(\cdot)$, its update uses kernel gradient descent and associates the same error
with the graph-based representation at which it occurred. Thus, an error
on a node/edge has greater influence on future nodes/edges with similar encoded representations.

Algorithm~\ref{alg:glocalcp} summarizes the complete online procedure.
%All quantities used to construct $\mathcal C_n$ are computed before $y_n$ is observed; the revealed label is used only for the subsequent calibration update.

We next establish that this uncertainty-adaptive, graph-localized procedure
retains pathwise marginal calibration even on evolving and dependent graph
streams.

\begin{theorem}[Pathwise marginal calibration on evolving graphs]
\label{thm:marginal}
Consider any evolving graph sequence
$\{\mathcal G_n\}_{n=1}^{N}$ and any corresponding sequence of inductive
node or edge queries. Suppose $r_n$ is computed before observing $y_n$
and satisfies $r_n\in[\gamma_r,1]$ for some $\gamma_r>0$. Then the
GLocalCP sets in Equation~\eqref{eq:glocal_set} satisfy
\[
\left|
\frac{1}{N}
\sum_{n=1}^{N}
\mathbb{I}\{y_n\notin\mathcal C_n ({\bf z}_n)\}
-\alpha
\right|
=
O(N^{-1/2}).
\]
Consequently, the realized marginal coverage converges to $1-\alpha$
even when node queries have overlapping neighborhoods or edge queries
share endpoints, without requiring exchangeability or stationarity.
\end{theorem}

Theorem~\ref{thm:marginal} requires only that the uncertainty scale be
predictable and bounded away from zero; its particular MI construction is
not required for the guarantee. The graph may evolve and queries may remain
dependent through shared nodes, endpoints, and neighborhoods. The explicit
finite-sample bound and the corresponding proofs are provided in Appendix~\ref{app:proofs}.

% The marginal guarantee does not specify where errors occur in graph-query
% representation space. The local state $g_n$, however, is designed precisely
% to retain this information. Section~\ref{sec:local} shows that the same
% update further controls kernel-localized calibration over the inductive graph
% representations.

\section{Beyond Marginal Coverage: Graph-Localized Calibration}
\label{sec:local}

Theorem~\ref{thm:marginal} guarantees marginal calibration on evolving
graph streams, but does not constrain how miscoverage is distributed over
the graph-query representation space. The local update in
Equation~\eqref{eq:glocal_update} retains this information by associating
each realized calibration error with the representation ${\bf z}_n$ at which it
occurs. We now formalize graph-localized calibration and show that the same
feedback yields a pathwise guarantee.

For any representation ${\bf z}$, define the kernel-local miscoverage rate as
$\widehat R_{N,k}({\bf z})
=
\sum_{n=1}^{N}k({\bf z},{\bf z}_n)\mathbb{I}\{y_n\notin\mathcal C_n({\bf z}_n)\}
/\sum_{n=1}^{N}k({\bf z},{\bf z}_n)$.
We define the corresponding local calibration gap as
\begin{equation}
E_N({\bf z})
=
\left|
\widehat R_{N,k}({\bf z})-\alpha
\right|.
\label{eq:local_gap}
\end{equation}
Graph-localized calibration requires
$\sup_{\|{\bf z}\|_2\le1}E_N({\bf z})$ to be small over the graph-query representation
space. For node classification, ${\bf z}$ represents an inductive node context,
so $E_N({\bf z})$ measures calibration among nodes with similar representations;
for edge prediction, ${\bf z}$ represents a candidate-pair context and measures
calibration among edges with similar pair representations.

\begin{algorithm}[t]
\caption{GLocalCP on an Inductive Graph Stream}
\label{alg:glocalcp}
\begin{algorithmic}[1]
\Require Trained inductive encoder ${\bf g}_\theta$,
    node/edge prediction head, target miscoverage $\alpha$,
    kernel $k$, step size $\eta$, MC samples $M$,
    uncertainty floor $\gamma_r$
\State Initialize global state $q_1$ and graph-localized state $g_1=0$

\For{$n=1,2,\ldots$}
    \State Observe the current graph $\mathcal{G}_n$
 \Statex \textbf{S1. Graph-based prediction}
    \If{Node classification task with new node $v_n$}
        \State Obtain node embedding ${\bf z}_n (v_n)$ based on~(\ref{eq:node_embedding});
        \State Obtain class probabilities $\tilde{\bf p}_n(v_n)$ based on~(\ref{eq:node_prob});
        \State ${\bf z}_n (v_n) \rightarrow {\bf z}_n$ and $\tilde{\bf p}_n(v_n) \rightarrow \tilde{\bf p}_n$;
    \Else
        \State For $e_n=(v,v')$, obtain the edge embedding~(\ref{eq:edge_embedding}) from the node embeddings for $v$ and $v'$;
        \State Obtain probabilities $\tilde{\bf p}_n(e_n)$ based on~(\ref{eq:edge_prob});
        \State ${\bf z}_n (e_n) \rightarrow {\bf z}_n$ and $\tilde{\bf p}_n(e_n) \rightarrow \tilde{\bf p}_n$;
    \EndIf

    \Statex \textbf{S2. Uncertainty-adaptive score}
    \State Obtain $M$  predictions  $\{\widetilde{\mathbf{p}}_n^{(m)}\}_{m=1}^M$ using MC dropout and compute the average
        $\widetilde{\mathbf{p}}_n$ via~\eqref{eq:av};
    \State Compute predictive MI and the uncertainty scale $r_n$
        using Eqs.~\eqref{eq:predictive_mi}-\eqref{eq:mi_scale};
    \State Evaluate the graph-localized threshold $g_n({\bf z}_n)$;
    \State Construct conformal prediction set $\mathcal{C}_n$ using Eq.~\eqref{eq:glocal_set};

    \Statex \textbf{S3. Global and local thresholds update}
    \State Observe the node class or edge label $y_n$;
    \State Compute $\epsilon_n$ based on~\eqref{eq:coverage_error};
    \State Update the global $q_{n+1}$ and
        $g_{n+1}$ thresholds based on~\eqref{eq:glocal_update};
\EndFor

\end{algorithmic}
\end{algorithm}

For the exact-kernel analysis, assume that $k$ is fixed, symmetric, and
positive definite, with $k({\bf z},{\bf z})=1$ and
$k({\bf z},{\bf z}')\ge k_{\min}>0$ over the normalized graph-query representation
domain. The normalization bounds the RKHS feature norm, while uniform
positivity ensures that every anchor receives nonvanishing kernel mass.
Let $B_S:=1/\gamma_r$ and
$M_\alpha:=\max\{\alpha,1-\alpha\}$.

\begin{theorem}[Pathwise graph-localized calibration]
\label{thm:local}
Consider any evolving graph sequence and its corresponding inductive node
or edge representations $\{{\bf z}_n\}_{n=1}^{N}$ with $\|{\bf z}_n\|_2\le1$.
Under the kernel conditions above, the GLocalCP update in
Equation~\eqref{eq:glocal_update} satisfies
\begin{equation}
\sup_{\|z\|_2\le1} E_N({\bf z})
\le
\frac{
\sqrt{
2\eta(1-\alpha)B_S
+
2\eta^2M_\alpha^2
}
}{
\eta k_{\min}\sqrt N
}.
\label{eq:local_bound}
\end{equation}
Consequently,
$\sup_{\|{\bf z}\|_2\le1}E_N({\bf z})=O(N^{-1/2})$.
\end{theorem}
The proof of Theorem~\ref{thm:local} is provided in Appendix~\ref{app:proofs}.
Theorem~\ref{thm:local} complements the marginal guarantee in
Theorem~\ref{thm:marginal}. The global state $q_n$ controls aggregate
miscoverage, while the kernel state $g_n$ controls how that error is
distributed over the inductive graph-query representation space.
For node classification, the resulting guarantee is local to similar
encoded node contexts; for edge prediction, it is local to similar
candidate-pair contexts. Both results are pathwise and therefore allow
dependencies induced by overlapping neighborhoods, shared endpoints, and
an evolving graph, without requiring exchangeability or stationarity.

The guarantee is local at the resolution induced by the fixed kernel and
is not equivalent to per-query, class-conditional, or arbitrary
subgroup coverage. It applies to the exact-kernel update in
Equation~\eqref{eq:glocal_update}. For scalability, our large-graph
implementation approximates the local correction using random Fourier
features (RFF), which has been evaluated empirically in the ensuing section.

% Requires booktabs. Replace the original Experiments section only.
\section{Experiments}
\label{sec:experiments}

We evaluate GLocalCP to address three questions.
{\bf Q1.} Does GLocalCP maintain target marginal coverage while reducing
graph-local calibration error on inductive graph streams?
{\bf Q2.} Can similar marginal coverage conceal substantially different
graph-localized calibration behavior over time?
{\bf Q3.} How do local feedback and uncertainty scaling contribute to
calibration and efficiency, and do these effects persist across predictive
architectures, nonconformity scores, and edge-query distributions?

\subsection{Experimental Setup}
\label{sec:experimental_setup}
\vspace{-0.1cm}
We consider multiclass node classification and binary held-out edge prediction
on seven graph benchmarks, namely, Cora and CiteSeer~\citep{sen2008collective},
PubMed~\citep{namata2012query},
Computers~\citep{shchur2018pitfalls},
Roman-empire and Amazon-ratings~\citep{platonov2023heterophily}, and
ogbn-arxiv~\citep{hu2020ogb}. Nodes arrive sequentially in an inductive
setting. For node classification, approximately $10\%$ of the vertices form
the initial training graph and $2\%$ are used for validation. As for edge
prediction, the corresponding fractions are $20\%$ and $2\%$. The GNN is
trained only on the initial graph and remains frozen throughout the stream,
with each prediction using only information available at its arrival time.
Nodes in ogbn-arxiv arrive in publication-year order, with random ordering
within each year, whereas the remaining benchmarks use seeded random arrival
orders. The first $200$ queries form the calibration prefix; thereafter, each
label is revealed only after prediction.

Our default GLocalCP configuration uses GraphSAGE, MC-dropout prediction,
LAC, and MI-based uncertainty scaling. We compare against NAPS-U, DAPS, and
SNAPS for node classification, and CLP-CQR, CLP-SCQR, RR-CQR, and
RR-Mondrian for edge prediction, with the edge baselines retaining their
native CQR-based constructions. At target coverage $1-\alpha=0.9$, we report
marginal coverage (Cov.), mean prediction-set size (Size), and denote the
worst-case kernel-local calibration gap
$\sup_{\|z\|_2\le1}E_N(z)$ from Section~\ref{sec:local} by $K$.
All results are averaged over $50$ seeds. Additional details on stream
construction, edge-candidate generation, baseline adaptations,
hyperparameters, metric computation, and standard deviations are provided
in the appendix.

\subsection{Marginal and Graph-Localized Calibration}
\label{sec:main_results}

Table~\ref{tab:main_results} reports the primary node classification and edge prediction results.
GLocalCP maintains marginal coverage near the target $0.9$ across all seven
node and seven edge streams, in line with the pathwise marginal calibration
guarantee in Theorem~\ref{thm:marginal}. The node classification baselines also illustrate
why coverage and set size must be interpreted jointly. NAPS-U is strongly
conservative, often approaching unit coverage with large sets, whereas DAPS
and SNAPS frequently produce smaller sets at the cost of undercoverage. On
CiteSeer, for example, DAPS and SNAPS attain coverage $0.734$ and $0.767$,
respectively, compared with $0.900$ for GLocalCP.

More importantly, similar marginal coverage can coexist with substantially
different graph-localized calibration. This is especially clear for edge
prediction, where several baselines attain aggregate coverage near $0.9$ yet
retain much larger local gaps. On CiteSeer, for example, RR-Mondrian attains
coverage $0.911$, compared with $0.900$ for GLocalCP, but has a local gap of
$0.0540$ versus $0.0017$. Across all benchmarks, GLocalCP attains the smallest
$K$ on every node and edge dataset while maintaining near-target marginal
coverage. These results show that aggregate coverage alone can conceal
substantial calibration disparities across local regions of the
graph-query representation space.

\begin{table*}[t]
\centering
\scriptsize
\setlength{\tabcolsep}{3pt}

\begin{tabular}{lcccccccccccc}
\toprule
\multicolumn{13}{c}{\textbf{Node Classification}} \\
\midrule
 & \multicolumn{3}{c}{NAPS-U}
 & \multicolumn{3}{c}{DAPS}
 & \multicolumn{3}{c}{SNAPS}
 & \multicolumn{3}{c}{GLocalCP} \\
Dataset
 & Cov. & Size & $K$
 & Cov. & Size & $K$
 & Cov. & Size & $K$
 & Cov. & Size & $K$ \\
\midrule
Cora
 & 0.999 & 6.822 & 0.0991
 & 0.766 & 1.925 & 0.1848
 & 0.783 & 1.808 & 0.1683
 & \textbf{0.902} & \textbf{2.527} & \textbf{0.0089} \\
CiteSeer
 & 1.000 & 5.980 & 0.0999
 & 0.734 & 2.148 & 0.1984
 & 0.767 & 2.033 & 0.1619
 & \textbf{0.900} & \textbf{2.972} & \textbf{0.0050} \\
PubMed
 & 1.000 & 2.999 & 0.1000
 & 0.830 & 1.031 & 0.1088
 & 0.852 & 1.042 & 0.0809
 & \textbf{0.900} & \textbf{1.156} & \textbf{0.0055} \\
Computers
 & 0.966 & 6.002 & 0.0819
 & 0.873 & 1.536 & 0.1191
 & 0.882 & 1.544 & 0.1082
 & \textbf{0.900} & \textbf{1.473} & \textbf{0.0139} \\
Roman-empire
 & 1.000 & 18.000 & 0.1000
 & 0.764 & 5.399 & 0.2222
 & 0.762 & 4.904 & 0.2355
 & \textbf{0.900} & \textbf{5.363} & \textbf{0.0179} \\
Amazon-ratings
 & 1.000 & 5.000 & 0.1000
 & 0.831 & 2.815 & 0.0826
 & 0.828 & 2.809 & 0.0858
 & \textbf{0.900} & \textbf{3.347} & \textbf{0.0007} \\
ogbn-arxiv
 & 1.000 & 40.000 & 0.1000
 & 0.865 & 6.221 & 0.0541
 & 0.861 & 6.093 & 0.0584
 & \textbf{0.900} & \textbf{10.563} & \textbf{0.0028} \\
\bottomrule
\end{tabular}

\vspace{5pt}

\begin{tabular}{lccccccccccccccc}
\toprule
\multicolumn{16}{c}{\textbf{Held-out Edge Prediction ($1{:}10$)}} \\
\midrule
 & \multicolumn{3}{c}{CLP-CQR}
 & \multicolumn{3}{c}{CLP-SCQR}
 & \multicolumn{3}{c}{RR-CQR}
 & \multicolumn{3}{c}{RR-Mondrian}
 & \multicolumn{3}{c}{GLocalCP} \\
Dataset
 & Cov. & Size & $K$
 & Cov. & Size & $K$
 & Cov. & Size & $K$
 & Cov. & Size & $K$
 & Cov. & Size & $K$ \\
\midrule
Cora
 & 0.897 & 1.815 & 0.0852
 & 0.901 & 1.818 & 0.0839
 & 0.906 & \textbf{1.361} & 0.0433
 & 0.905 & 1.508 & 0.0432
 & \textbf{0.900} & 1.559 & \textbf{0.0009} \\
CiteSeer
 & 0.908 & 1.784 & 0.0714
 & 0.911 & 1.784 & 0.0728
 & 0.909 & \textbf{1.427} & 0.0559
 & 0.911 & 1.520 & 0.0540
 & \textbf{0.900} & 1.581 & \textbf{0.0017} \\
PubMed
 & 0.894 & 1.620 & 0.0710
 & 0.891 & 1.595 & 0.0748
 & 0.888 & 1.720 & 0.0793
 & 0.873 & 1.723 & 0.0859
 & \textbf{0.900} & \textbf{1.454} & \textbf{0.0003} \\
Computers
 & 0.904 & 1.755 & 0.0528
 & 0.908 & 1.715 & 0.0525
 & 0.943 & 1.805 & 0.0745
 & 0.924 & 1.860 & 0.0707
 & \textbf{0.900} & \textbf{1.372} & \textbf{0.0040} \\
Roman-empire
 & 0.887 & 1.799 & 0.0660
 & 0.887 & 1.799 & 0.0681
 & 0.892 & 1.404 & 0.0352
 & 0.918 & 1.602 & 0.0427
 & \textbf{0.900} & \textbf{1.542} & \textbf{0.0007} \\
Amazon-ratings
 & 0.893 & 1.846 & 0.0629
 & 0.894 & 1.831 & 0.0672
 & 0.902 & 1.635 & 0.0513
 & 0.908 & 1.688 & 0.0507
 & \textbf{0.900} & \textbf{1.437} & \textbf{0.0007} \\
ogbn-arxiv
 & 0.882 & 1.774 & 0.1001
 & 0.883 & 1.713 & 0.0977
 & 0.940 & 1.852 & 0.0906
 & 0.920 & 1.842 & 0.0664
 & \textbf{0.900} & \textbf{1.331} & \textbf{0.0021} \\
\bottomrule
\end{tabular}
\caption{Node and edge results at $\alpha=0.1$.
Bold marks coverage closest to $0.9$, the smallest Size among methods with
coverage at least $0.9$, and the smallest $K$.}
\label{tab:main_results}
\end{table*}

\begin{figure*}[t]
    \centering
    \includegraphics[
        width=\textwidth,
        trim={0cm 1cm 0cm 0cm},
        clip
    ]{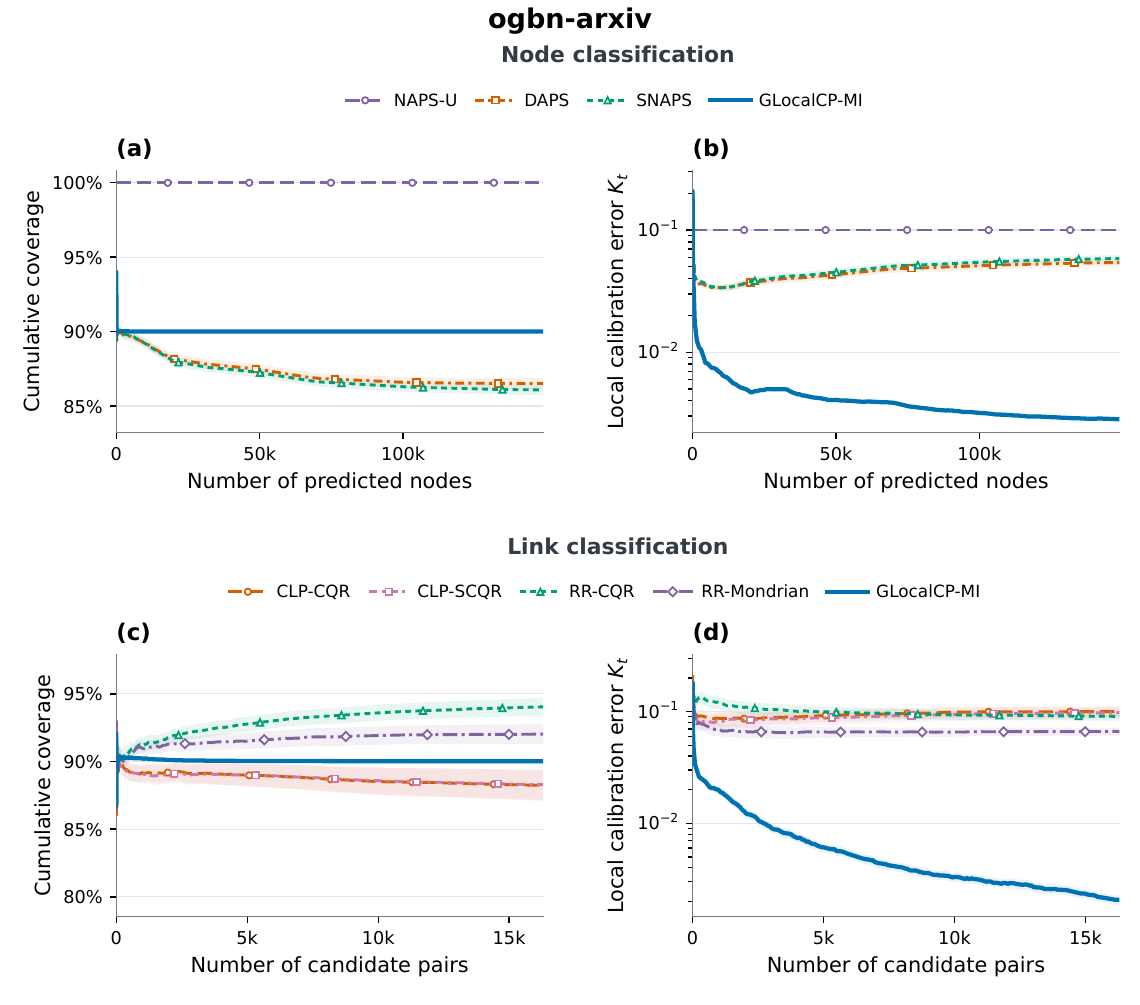}
    \caption{Stream-wise marginal and graph-local calibration on ogbn-arxiv
    at target coverage $1-\alpha=0.9$. (a,c) Cumulative marginal coverage for
    node and held-out edge prediction. (b,d) Prefix-wise graph-local
    calibration gap $K_t$ on a logarithmic scale. GLocalCP maintains marginal
    coverage near the target while progressively reducing graph-local
    calibration error.}
    \label{fig:stream_dynamics}
\end{figure*}

\textbf{Calibration dynamics.}
Figure~\ref{fig:stream_dynamics} examines how this distinction evolves along
the ogbn-arxiv stream. For both node classification and edge prediction, cumulative marginal
coverage makes several methods appear increasingly similar as the stream
progresses. Their graph-localized trajectories reveal a different pattern.
Let $K_t$ denote the prefix-wise counterpart of $K$ computed from the first
$t$ queries. GLocalCP reduces $K_t$ toward the $10^{-3}$ scale, whereas the
competing methods retain substantially larger local gaps even when their
aggregate coverage is close to the target. This separation persists beyond
initialization and illustrates the distinction in Section~\ref{sec:local}:
marginal calibration does not determine how coverage errors are distributed
across the graph-query representation space. Corresponding stream-wise plots
for the remaining datasets are provided in the appendix.

\subsection{Ablation and Robustness Studies}
\label{sec:ablation_robustness}
\vspace{-0.2cm}

We first isolate the two mechanisms in GLocalCP through controlled
comparisons. To assess local feedback, we compare GLocalCP-MI with Global MI,
which uses the same MI-dependent uncertainty scale but sets $g_n\equiv0$.
Adding local feedback reduces $K$ on all seven node and seven edge datasets.
For example, node $K$ decreases from $0.0621$ to $0.0139$ on Computers and
from $0.0145$ to $0.0028$ on ogbn-arxiv; edge $K$ decreases from $0.0573$
to $0.0017$ on CiteSeer and from $0.0210$ to $0.0003$ on PubMed. Thus, with
uncertainty scaling fixed, the kernel-local correction drives the reduction
in graph-local calibration error.

To isolate uncertainty scaling, we compare GLocalCP-MI with Local $r=1$,
which retains the same local feedback but removes the MI-dependent scale.
Marginal coverage remains essentially unchanged, while MI scaling reduces
mean prediction-set size on all seven node and seven edge datasets. On
ogbn-arxiv nodes, Size decreases from $13.879$ to $10.563$; on CiteSeer
edges, it decreases from $1.650$ to $1.581$. Together, these comparisons
separate the roles of the two components: kernel-local feedback improves
graph-local calibration, whereas MI scaling primarily improves
prediction-set efficiency.

\begin{table*}[t]
\centering
\small
\setlength{\tabcolsep}{3pt}
\renewcommand{\arraystretch}{1.08}

\begin{tabular}{@{}lrrrrrrrr@{}}
\toprule
& \multicolumn{4}{c}{(a) Effect of local feedback}
& \multicolumn{4}{c}{(b) Effect of MI scaling} \\
\cmidrule(lr){2-5}
\cmidrule(lr){6-9}
& \multicolumn{2}{c}{Node $K$}
& \multicolumn{2}{c}{Edge $K$}
& \multicolumn{2}{c}{Node Size}
& \multicolumn{2}{c}{Edge Size} \\
\cmidrule(lr){2-3}
\cmidrule(lr){4-5}
\cmidrule(lr){6-7}
\cmidrule(lr){8-9}
Dataset
& Global MI & GLocalCP
& Global MI & GLocalCP
& $r_n=1$ & MI
& $r_n=1$ & MI \\
\midrule
Cora
& 0.0159 & \textbf{0.0089}
& 0.0386 & \textbf{0.0009}
& 2.631 & \textbf{2.527}
& 1.643 & \textbf{1.559} \\
CiteSeer
& 0.0121 & \textbf{0.0050}
& 0.0573 & \textbf{0.0017}
& 3.081 & \textbf{2.972}
& 1.650 & \textbf{1.581} \\
PubMed
& 0.0209 & \textbf{0.0055}
& 0.0210 & \textbf{0.0003}
& 1.189 & \textbf{1.156}
& 1.551 & \textbf{1.454} \\
Computers
& 0.0621 & \textbf{0.0139}
& 0.0453 & \textbf{0.0040}
& 1.597 & \textbf{1.473}
& 1.396 & \textbf{1.372} \\
Roman-empire
& 0.0471 & \textbf{0.0179}
& 0.0261 & \textbf{0.0007}
& 6.242 & \textbf{5.363}
& 1.641 & \textbf{1.542} \\
Amazon-ratings
& 0.0062 & \textbf{0.0007}
& 0.0147 & \textbf{0.0007}
& 3.480 & \textbf{3.347}
& 1.536 & \textbf{1.437} \\
ogbn-arxiv
& 0.0145 & \textbf{0.0028}
& 0.0247 & \textbf{0.0021}
& 13.879 & \textbf{10.563}
& 1.389 & \textbf{1.331} \\
\bottomrule
\end{tabular}

\caption{Controlled ablations of the two GLocalCP components at
$\alpha=0.1$. (a) Effect of local feedback: Global MI and GLocalCP use the
same MI-dependent scale, but Global MI sets $g_n\equiv0$. Adding local
feedback reduces $K$ on all seven node and seven edge datasets.
(b) Effect of uncertainty scaling: both variants use the same kernel-local
feedback, with either $r_n=1$ or MI-based scaling. MI scaling reduces mean
prediction-set size on all seven node and seven edge datasets while
preserving near-target marginal coverage. Values are means over $50$ seeds.
Bold indicates the smaller value within each paired comparison.}
\label{tab:component_ablation}
\end{table*}

We next test whether these effects persist across predictive architectures
and nonconformity scores. Replacing GraphSAGE with GCN and LAC with APS
preserves the main pattern: GLocalCP maintains near-target marginal coverage
and small graph-localized gaps, although prediction-set size varies across
configurations. Complete results for GCN--LAC, GraphSAGE--APS, and GCN--APS
on all seven node and edge datasets are reported in
Appendix~\ref{app:updated_experiment_results}.

\begin{wraptable}{r}{0.5\textwidth}
\centering
\small
\setlength{\tabcolsep}{7pt}
\renewcommand{\arraystretch}{1.08}
\caption{
Relative reduction (\%) in graph-localized gap $K$ of GLocalCP versus the
native edge baseline with the smallest $K$ under each candidate distribution.
Values are computed from $50$-seed mean gaps.
}
\label{tab:candidate_robustness}
\begin{tabular}{lrrr}
\toprule
Dataset & $1{:}10$ & $1{:}1$ & Hard \\
\midrule
Cora           & 97.8 & 91.9 & 89.7 \\
CiteSeer       & 96.8 & 89.7 & 88.6 \\
PubMed         & 99.6 & 97.0 & 95.6 \\
Computers      & 92.4 & 76.4 & 59.3 \\
Roman-empire   & 98.0 & 95.6 & 94.9 \\
Amazon-ratings & 98.7 & 92.6 & 91.0 \\
ogbn-arxiv     & 96.9 & 84.5 & 84.0 \\
\bottomrule
\end{tabular}
\end{wraptable}
Finally, we vary the edge-query distribution. In addition to the standard
$1{:}10$ candidate ratio, we evaluate balanced $1{:}1$ candidates and hard
two-hop negatives. Table~\ref{tab:candidate_robustness} reports the relative
reduction in $K$ of GLocalCP against the native edge baseline with the
smallest $K$ in each setting. The reduction is positive across all $21$
dataset--candidate combinations, ranging from $76.4\%$ to $97.0\%$ under
balanced sampling and from $59.3\%$ to $95.6\%$ under hard negatives.
Although the margin narrows on some harder streams, most notably Computers,
the graph-localized calibration improvement persists across all seven datasets.

Overall, the controlled studies show that kernel-localized feedback improves graph-localized calibration and MI-based uncertainty scaling improves prediction-set efficiency. These observations are consistent across the evaluated models and prediction tasks (node classification and edge prediction).

\section{Conclusion}
\label{sec:conclusion}
\vspace{-0.2cm}
This paper proposed  GLocalCP for inductive graph predictions beyond marginal
coverage. Specifically, GLocalCP
uses predictive uncertainty from MC dropout to adapt prediction-set allocation and inductive
graph representations to share calibration feedback among similar node or
edge queries. Theoretically, we established pathwise $O(N^{-1/2})$ marginal and
 graph-localized calibration guarantees without any assumption of exchangeability or
stationarity. Across node classification and edge prediction benchmarks,
GLocalCP maintained near-target marginal coverage while consistently
reducing graph-localized calibration error, and uncertainty scaling further
improved prediction-set efficiency. These results highlight graph-localized
calibration as an important complement to marginal validity for uncertainty
quantification on evolving graphs.

\clearpage

\section*{Reproducibility Statement}
Complete proofs and experimental details are provided in the appendix to
support reproduction of the theoretical and empirical results.

\section*{Ethics Statement}
This study used public datasets and needs no ethical approval.

\section*{AI Use Statement}
Generative AI was used only to assist with writing and language polishing.

\bibliographystyle{iclr2027_conference}
\bibliography{references}
\clearpage

\appendix
\begin{center}
    {\Large\bf Supplementary File}
\end{center}
\section{Related Work}
\label{sec:related_work}

\paragraph{Conformal prediction (CP) on graphs.}
A growing literature extends CP to graph-structured
data, where observations are dependent and graph structure can itself inform
uncertainty. Early work studies conformal inference with network-derived
covariates and establishes validity under suitable network exchangeability
conditions~\citep{lunde2023conformal}. For GNN-based node classification,
DAPS~\citep{zargarbashi2023conformal} diffuses conformity scores over graph
neighborhoods to exploit homophily, while
SNAPS~\citep{song2024snaps} combines structural neighborhoods with feature
similarity to improve prediction-set efficiency. CF-GNN
\citep{huang2023uncertainty} establishes permutation-invariance
conditions for conformal validity on graphs and learns a topology-aware
output correction to improve efficiency. These methods primarily study
prediction on an observed graph. NAPS~\citep{clarkson2023naps}, in contrast,
considers inductive node classification and locally reweights calibration
scores according to the neighborhood of a newly observed node. Subsequent
work further develops CP for inductive GNNs while accounting for changes in
node representations when new nodes enter the graph
\citep{zargarbashi2024conformalinductive}. Other recent approaches improve
graph conformal efficiency through graph-aware training or structural
modification, rather than calibration alone~\citep{he2024spargcp}.

Graph CP has also expanded beyond node classification.
CLP~\citep{zhao2024clp} develops conformalized link prediction and exploits
graph-specific edge constructions to improve efficiency, while
RR-GNN~\citep{zhang2025rr} combines residual-adaptive scores with
graph-structured Mondrian calibration to account for topological and
task-dependent heterogeneity. Together, these works establish several ways
to exploit graph structure and, in some cases, accommodate previously unseen
nodes or edges. Inductive capability alone, however, does not imply
adaptation to distribution shift along a sequential graph stream. As nodes
and edge queries arrive, their representations, observed neighborhoods, and
query distribution can evolve, so calibration relationships appropriate at
one stage need not persist throughout the stream. Our focus is therefore
different: we study feedback-driven calibration along an evolving inductive
graph stream and explicitly track how realized coverage errors vary over the
graph-query representation space.

\paragraph{CP under shift and sequential feedback.}
Outside the graph setting, substantial work has relaxed the classical
exchangeability assumption. Weighted CP handles covariate shift by
reweighting calibration observations according to the train–test covariate
distribution~\citep{tibshirani2019conformal}, while conformal prediction
beyond exchangeability quantifies validity under more general departures
from exchangeability~\citep{barber2023beyond}. Adaptive conformal inference
(ACI) instead uses sequential coverage feedback to update calibration under
distribution shift~\citep{gibbs2021aci}. Follow-up methods improve
adaptation to changing environments through dynamically tuned learning rates
and expert aggregation~\citep{gibbs2024conformal}, strongly adaptive online
learning~\citep{bhatnagar2023strongly}, and feedback-control formulations
such as conformal PID~\citep{angelopoulos2024pid}. These developments show
that useful long-run or interval-wise calibration guarantees can be obtained
even when the data distribution changes over time.

A related literature addresses heterogeneity across the covariate space.
Exact distribution-free conditional coverage is generally unattainable
without additional assumptions, motivating weaker or localized notions of
conditional validity~\citep{lei2014distribution,vovk2013conditional}.
Localized CP uses test-dependent weighting to adapt conformal inference to
nearby covariates while retaining marginal validity and, under additional
conditions, obtaining local coverage properties~\citep{guan2023localized}.
More recent Online Localized Conformal Prediction extends localization to
sequential prediction and adapts across heterogeneous covariate regions
\citep{lai2026online}. These methods provide complementary tools for
distribution shift, temporal adaptation, and covariate heterogeneity; they
are not specifically designed around evolving graph-structured queries or
the graph-local reliability objective considered here.

GLocalCP connects these lines of work from the perspective of graph
prediction. It applies feedback-driven online calibration to sequential node
and edge queries, retains the representation at which each coverage error
occurs through a kernel-local state, and separately uses predictive
uncertainty to adapt prediction-set allocation. This yields pathwise
marginal and graph-local calibration guarantees tailored to evolving
inductive graph streams.
\section{Proofs}
\label{app:proofs}

\subsection{Proof Setup and Auxiliary Results}
\label{app:proof_notation}

We first introduce the common notation used to prove the marginal and
graph-localized guarantees. The analysis applies to either node or edge queries:
${\bf z}_n$ denotes the corresponding inductive graph-query representation,
$r_n$ its predictable uncertainty scale, and
$\mathcal C_n({\bf z}_n)$ the GLocalCP prediction set in
Equation~\eqref{eq:glocal_set}.

For the realized label $y_n$, define the uncertainty-scaled score
\[
S_n
=
\frac{s({\bf z}_n,y_n)}{r_n}.
\]
Since $s(y_n;\widetilde{\mathbf p}_n)\in[0,1]$ and
$r_n\ge\gamma_r$, it satisfies
\begin{equation}
0\le S_n\le B_S,
\qquad
B_S:=\frac{1}{\gamma_r}.
\label{eq:local_score_bound}
\end{equation}
By Equation~\eqref{eq:glocal_set}, miscoverage occurs exactly when
\begin{equation}
y_n\notin\mathcal C_n({\bf z}_n)
\quad\Longleftrightarrow\quad
S_n>q_n+g_n({\bf z}_n).
\label{eq:local_score_set_equiv}
\end{equation}

To analyze the global and local updates jointly, consider the direct-sum
Hilbert space
$\overline{\mathcal H}:=\mathbb R\oplus\mathcal H_k$ and define
\[
H_n:=(q_n,g_n),
\qquad
\Phi_n:=\bigl(1,k(\cdot,{\bf z}_n)\bigr).
\]
Then the threshold and update in
Equation~\eqref{eq:glocal_update} can be written compactly as
\begin{equation}
q_n+g_n({\bf z}_n)
=
\langle H_n,\Phi_n\rangle_{\overline{\mathcal H}},
\qquad
H_{n+1}
=
H_n+\eta \epsilon_n\Phi_n,
\label{eq:local_joint_update}
\end{equation}
where
\[
\epsilon_n
=
\mathbb{I}\{y_n\notin\mathcal C_n({\bf z}_n)\}-\alpha.
\]
For the explicit bounds below, we take the zero-state initialization
$H_1=(0,0)$.

\begin{lemma}[Uniform kernel lower bound]
\label{lem:kernel_lower_bound}
Suppose the normalized graph-query representations satisfy
$\|{\bf z}\|_2\le1$. The RBF and Mat\'ern kernels defined in
\Cref{app:kernel_constructions} satisfy $k({\bf z},{\bf z})=1$ and admit the
uniform lower bounds
\begin{equation}
k_{\min}
=
e^{-2/h^2}
\quad\text{for RBF},
\qquad
k_{\min}
=
k_{\nu,h}^{\rm Mat}(2)
\quad\text{for Mat\'ern}.
\label{eq:local_kernel_lower_bound}
\end{equation}
Consequently,
\begin{equation}
\sum_{n=1}^{N}k({\bf z},{\bf z}_n)
\ge
Nk_{\min}
\qquad
\text{for every }\|{\bf z}\|_2\le1.
\label{eq:local_kernel_mass}
\end{equation}
\end{lemma}

The next lemma provides the common potential bound underlying both
Theorem~\ref{thm:marginal} and Theorem~\ref{thm:local}.

\begin{lemma}[Residual-potential bound]
\label{lem:local_residual_potential}
Let $M_\alpha:=\max\{\alpha,1-\alpha\}$ and initialize $H_1=(0,0)$.
Then, for every realized sequence of inductive node or edge queries,
\begin{equation}
\|H_{N+1}\|_{\overline{\mathcal H}}^2
\le
N\left(
2\eta(1-\alpha)B_S
+
2\eta^2M_\alpha^2
\right).
\label{eq:local_potential_bound}
\end{equation}
\end{lemma}

The two main results follow by projecting this joint potential bound onto
its global and graph-localized components. The scalar component yields the
marginal calibration bound in Theorem~\ref{thm:marginal}, whereas the RKHS
component, together with Lemma~\ref{lem:kernel_lower_bound}, yields the
uniform graph-localized bound in Theorem~\ref{thm:local}.

\begin{proposition}[Invariance to predictable uncertainty scaling]
\label{prop:coverage_invariance}
The calibration guarantees require the uncertainty scales only to be
predictable and uniformly bounded as $r_n\in[\gamma_r,1]$. In particular,
the proof depends on $r_n$ only through the score bound in
Equation~\eqref{eq:local_score_bound} and the miscoverage equivalence in
Equation~\eqref{eq:local_score_set_equiv}; it does not require the specific
MI construction in Equation~\eqref{eq:mi_scale}.
\end{proposition}

\subsection{Proof of the Uniform Kernel Lower Bound}
\label{app:proof_kernel_lower}

\begin{proof}[Proof of Lemma~\ref{lem:kernel_lower_bound}]
Let
\[
\mathcal Z
=
\{{\bf z}\in\mathbb R^d:\|{\bf z}\|_2\le1\}
\]
denote the normalized graph-query representation domain. For any
${\bf z},{\bf z}'\in\mathcal Z$,
\[
\|{\bf z}-{\bf z}'\|_2
\le
\|{\bf z}\|_2+\|{\bf z}'\|_2
\le2.
\]
Thus, all pairs of inductive node or edge representations lie within
distance $2$.

For the RBF kernel,
\[
k_h^{\mathrm{RBF}}({\bf z},{\bf z}')
=
\exp\left(
-\frac{\|{\bf z}-{\bf z}'\|_2^2}{2h^2}
\right).
\]
Hence $k_h^{\mathrm{RBF}}({\bf z},{\bf z})=1$, and the distance bound above gives
\[
k_h^{\mathrm{RBF}}({\bf z},{\bf z}')
\ge
\exp\left(-\frac{2^2}{2h^2}\right)
=
e^{-2/h^2}.
\]
The lower bound is attained by antipodal unit vectors, so
\[
\inf_{{\bf z},{\bf z}'\in\mathcal Z}
k_h^{\mathrm{RBF}}({\bf z},{\bf z}')
=
e^{-2/h^2}.
\]

For the Mat\'ern kernel,
$k_{\nu,h}^{\mathrm{Mat}}({\bf z},{\bf z})=1$ by its continuous extension at
zero. Let
\[
x
=
\frac{\sqrt{2\nu}\,\|{\bf z}-{\bf z}'\|_2}{h}.
\]
Its radial profile is proportional to $x^\nu K_\nu(x)$. For $x>0$,
\[
\frac{d}{dx}
\left(x^\nu K_\nu(x)\right)
=
-x^\nu K_{\nu-1}(x)
<0,
\]
since $K_{\nu-1}(x)>0$. The Mat\'ern kernel is therefore decreasing in
$\|{\bf z}-{\bf z}'\|_2$. Since $\|{\bf z}-{\bf z}'\|_2\le2$,
\[
k_{\nu,h}^{\mathrm{Mat}}({\bf z},{\bf z}')
\ge
k_{\nu,h}^{\mathrm{Mat}}(2)>0,
\]
with equality attainable for antipodal unit vectors. Therefore,
\[
\inf_{{\bf z},{\bf z}'\in\mathcal Z}
k_{\nu,h}^{\mathrm{Mat}}({\bf z},{\bf z}')
=
k_{\nu,h}^{\mathrm{Mat}}(2).
\]

Thus both kernels are normalized and uniformly positive on the graph-query
representation domain, with the constants stated in
Equation~\eqref{eq:local_kernel_lower_bound}. Consequently,
\[
\sum_{n=1}^{N}k({\bf z},{\bf z}_n)
\ge
Nk_{\min}
\]
uniformly over ${\bf z}\in\mathcal Z$.
\end{proof}

\subsection{Proof of the Residual-Potential Bound}
\label{app:proof_potential}

\begin{proof}[Proof of Lemma~\ref{lem:local_residual_potential}]
Recall from Equation~\eqref{eq:local_joint_update} that
\[
t_n
=
q_n+g_n({\bf z}_n)
=
\langle H_n,\Phi_n\rangle_{\overline{\mathcal H}},
\qquad
H_{n+1}
=
H_n+\eta\epsilon_n\Phi_n.
\]
The key step is to control the interaction term $\epsilon_nt_n$ using the
prediction event.

If $S_n>t_n$, then
Equation~\eqref{eq:local_score_set_equiv} implies
$y_n\notin\mathcal C_n({\bf z}_n)$, so $\epsilon_n=1-\alpha$. Moreover,
$S_n\le B_S$ gives $t_n<B_S$, and hence
\[
\epsilon_nt_n
=
(1-\alpha)t_n
\le
(1-\alpha)B_S.
\]
If $S_n\le t_n$, then
$y_n\in\mathcal C_n({\bf z}_n)$ and $\epsilon_n=-\alpha$.
Since $S_n\ge0$, we have $t_n\ge0$, and therefore
\[
\epsilon_nt_n
=
-\alpha t_n
\le0.
\]
Thus, in either case,
\begin{equation}
\epsilon_nt_n
\le
(1-\alpha)B_S.
\label{eq:local_sign_control}
\end{equation}

Next, by the direct-sum norm and the reproducing property,
\[
\|\Phi_n\|_{\overline{\mathcal H}}^2
=
1+\|k(\cdot,{\bf z}_n)\|_{\mathcal H_k}^2
=
1+k({\bf z}_n,{\bf z}_n)
=
2,
\]
where the last equality follows from the kernel normalization
$k({\bf z}_n,{\bf z}_n)=1$. Also,
$|\epsilon_n|\le M_\alpha$, and hence $\epsilon_n^2\le M_\alpha^2$.

Expanding one step of the joint update gives
\begin{align*}
\|H_{n+1}\|_{\overline{\mathcal H}}^2
&=
\|H_n+\eta\epsilon_n\Phi_n\|_{\overline{\mathcal H}}^2\\
&=
\|H_n\|_{\overline{\mathcal H}}^2
+
2\eta\epsilon_n
\langle H_n,\Phi_n\rangle_{\overline{\mathcal H}}
+
\eta^2\epsilon_n^2
\|\Phi_n\|_{\overline{\mathcal H}}^2\\
&=
\|H_n\|_{\overline{\mathcal H}}^2
+
2\eta\epsilon_nt_n
+
\eta^2\epsilon_n^2
\|\Phi_n\|_{\overline{\mathcal H}}^2\\
&\le
\|H_n\|_{\overline{\mathcal H}}^2
+
2\eta(1-\alpha)B_S
+
2\eta^2M_\alpha^2.
\end{align*}
Summing over $n=1,\ldots,N$ and using $H_1=(0,0)$ yields
\[
\|H_{N+1}\|_{\overline{\mathcal H}}^2
\le
N\left(
2\eta(1-\alpha)B_S
+
2\eta^2M_\alpha^2
\right),
\]
which is exactly Equation~\eqref{eq:local_potential_bound}.
\end{proof}

\subsection{Proof of Pathwise Marginal Calibration}
\label{app:proof_marginal}

\begin{proof}[Proof of Theorem~\ref{thm:marginal}]
Recall the centered miscoverage feedback
\[
\epsilon_n
=
\mathbb{I}\{y_n\notin\mathcal C_n({\bf z}_n)\}-\alpha.
\]
The global component of the joint update in
Equation~(\ref{eq:glocal_update}) is
\[
q_{n+1}
=
q_n+\eta\epsilon_n.
\]
Summing this recursion from $n=1$ to $N$ gives
\begin{align*}
q_{N+1}-q_1
&=
\eta\sum_{n=1}^{N}\epsilon_n =
\eta\sum_{n=1}^{N}
\left(
\mathbb{I}\{y_n\notin\mathcal C_n({\bf z}_n)\}-\alpha
\right).
\end{align*}
Under the zero-state initialization $q_1=0$,
\[
\frac{1}{N}
\sum_{n=1}^{N}
\mathbb{I}\{y_n\notin\mathcal C_n({\bf z}_n)\}
-\alpha
=
\frac{q_{N+1}}{\eta N}.
\]
Thus, controlling empirical marginal miscoverage reduces to controlling
the growth of the global calibration state $q_{N+1}$.

The global state is one component of the joint state
$H_{N+1}=(q_{N+1},g_{N+1})\in
\mathbb R\oplus\mathcal H_k$. By the direct-sum norm,
\[
\|H_{N+1}\|_{\overline{\mathcal H}}^2
=
|q_{N+1}|^2
+
\|g_{N+1}\|_{\mathcal H_k}^2,
\]
and therefore
\[
|q_{N+1}|
\le
\|H_{N+1}\|_{\overline{\mathcal H}}.
\]
Lemma~\ref{lem:local_residual_potential} bounds this joint state as
\[
\|H_{N+1}\|_{\overline{\mathcal H}}^2
\le
N\left(
2\eta(1-\alpha)B_S
+
2\eta^2M_\alpha^2
\right).
\]
Taking square roots yields
\[
|q_{N+1}|
\le
\sqrt{N}
\sqrt{
2\eta(1-\alpha)B_S
+
2\eta^2M_\alpha^2
}.
\]
Substituting this bound into the telescoping identity above gives
\[
\begin{aligned}
\left|
\frac{1}{N}
\sum_{n=1}^{N}
\mathbb{I}\{y_n\notin\mathcal C_n({\bf z}_n)\}
-\alpha
\right|
&=
\frac{|q_{N+1}|}{\eta N} \le
\frac{
\sqrt{
2\eta(1-\alpha)B_S
+
2\eta^2M_\alpha^2
}
}{
\eta\sqrt N
}.
\end{aligned}
\]
Since $B_S=1/\gamma_r<\infty$ and $\eta>0$ are fixed, the right-hand
side is $O(N^{-1/2})$. Hence
\[
\frac{1}{N}
\sum_{n=1}^{N}
\mathbb{I}\{y_n\notin\mathcal C_n({\bf z}_n)\}
\longrightarrow
\alpha,
\]
or equivalently,
\[
\frac{1}{N}
\sum_{n=1}^{N}
\mathbb{I}\{y_n\in\mathcal C_n({\bf z}_n)\}
\longrightarrow
1-\alpha.
\]

It remains to clarify why this conclusion applies to the evolving graph
setting. The argument above is pathwise: after fixing any realized graph
sequence $\{\mathcal G_n\}_{n=1}^{N}$, the corresponding node or edge
queries, their inductive representations $\{{\bf z}_n\}_{n=1}^{N}$, and the
revealed labels $\{y_n\}_{n=1}^{N}$, every step is deterministic.
The proof uses only the predictability and boundedness of $r_n$ together
with the joint update in Equation~\eqref{eq:glocal_update}. It does not
factorize the joint distribution of the queries or labels, nor invoke
independence, exchangeability, or stationarity. Consequently, the same
bound holds when the graph evolves over time, node queries have overlapping
neighborhoods, or edge queries share endpoints.
\end{proof}

\subsection{Proof of Pathwise Graph-Localized Calibration}
\label{app:proof_local}

\begin{proof}[Proof of Theorem~\ref{thm:local}]
The RKHS component of the GLocalCP update satisfies
\[
g_{n+1}
=
g_n+\eta\epsilon_n k(\cdot,{\bf z}_n),
\qquad
\epsilon_n
=
\mathbb{I}\{y_n\notin\mathcal C_n({\bf z}_n)\}-\alpha.
\]
With $g_1=0$, summing over $n=1,\ldots,N$ gives
\[
g_{N+1}
=
\eta
\sum_{n=1}^{N}
\epsilon_n k(\cdot,{\bf z}_n).
\]
Evaluating this identity at any anchor ${\bf z}$ and using the reproducing
property yields
\[
g_{N+1}({\bf z})
=
\eta
\sum_{n=1}^{N}
k({\bf z},{\bf z}_n)
\left(
\mathbb{I}\{y_n\notin\mathcal C_n({\bf z}_n)\}-\alpha
\right).
\]
Therefore,
\[
\sum_{n=1}^{N}
k({\bf z},{\bf z}_n)
\left(
\mathbb{I}\{y_n\notin\mathcal C_n({\bf z}_n)\}-\alpha
\right)
=
\frac{g_{N+1}({\bf z})}{\eta}.
\]

By the definition of $E_N({\bf z})$ in Sec.~\ref{sec:local},
\begin{align*}
E_N({\bf z})
&=
\left|
\frac{
\sum_{n=1}^{N}
k({\bf z},{\bf z}_n)
\left(
\mathbb{I}\{y_n\notin\mathcal C_n({\bf z}_n)\}-\alpha
\right)
}{
\sum_{n=1}^{N}k({\bf z},{\bf z}_n)
}
\right| =
\frac{|g_{N+1}({\bf z})|}
{\eta\sum_{n=1}^{N}k({\bf z},{\bf z}_n)}.
\end{align*}

The reproducing property and Cauchy--Schwarz inequality give
\[
|g_{N+1}({\bf z})|
=
\left|
\left\langle
g_{N+1},k(\cdot,{\bf z})
\right\rangle_{\mathcal H_k}
\right|
\le
\|g_{N+1}\|_{\mathcal H_k}
\sqrt{k({\bf z},{\bf z})}
=
\|g_{N+1}\|_{\mathcal H_k},
\]
where $k({\bf z},{\bf z})=1$. Since $g_{N+1}$ is the RKHS component of
$H_{N+1}$,
\[
\|g_{N+1}\|_{\mathcal H_k}
\le
\|H_{N+1}\|_{\overline{\mathcal H}}.
\]
Lemma~\ref{lem:local_residual_potential} therefore implies
\[
|g_{N+1}({\bf z})|
\le
\sqrt{
N\left(
2\eta(1-\alpha)B_S
+
2\eta^2M_\alpha^2
\right)
}.
\]

Meanwhile, the uniform kernel lower bound in
Lemma~\ref{lem:kernel_lower_bound} gives
\[
\sum_{n=1}^{N}k({\bf z},{\bf z}_n)
\ge
Nk_{\min}
\]
for every normalized graph-query representation $\|{\bf z}\|_2\le1$.
Combining the numerator and denominator bounds yields
\[
E_N({\bf z})
\le
\frac{
\sqrt{
2\eta(1-\alpha)B_S
+
2\eta^2M_\alpha^2
}
}{
\eta k_{\min}\sqrt N
}
\]
uniformly over $\|{\bf z}\|_2\le1$. Taking the supremum over the
representation domain gives
\[
\sup_{\|{\bf z}\|_2\le1}E_N({\bf z})
\le
\frac{
\sqrt{
2\eta(1-\alpha)B_S
+
2\eta^2M_\alpha^2
}
}{
\eta k_{\min}\sqrt N
},
\]
which is Equation~\eqref{eq:local_bound}. Hence
\[
\sup_{\|{\bf z}\|_2\le1}E_N({\bf z})
=
O(N^{-1/2}).
\]

The argument is again pathwise. For node classification, ${\bf z}_n$ is the
inductive representation of the arriving node, so the bound controls
kernel-weighted calibration over similar encoded node contexts. For edge
prediction, ${\bf z}_n$ is the symmetric pair representation, so the same
bound controls calibration over similar candidate-edge contexts. No
independence, exchangeability, or stationarity assumption on these graph
queries is required.
\end{proof}

\subsection{Proof of Invariance to Predictable Uncertainty Scaling}
\label{app:proof_scaling_invariance}

\begin{proof}[Proof of Proposition~\ref{prop:coverage_invariance}]
The uncertainty scale enters GLocalCP only through the scaled
nonconformity score
\[
\frac{s(y_n;\widetilde{\mathbf p}_n)}{r_n}.
\]
Because $r_n$ is computed before observing $y_n$, the prediction set in
Equation~\eqref{eq:glocal_set} remains predictable. Moreover,
$r_n\ge\gamma_r>0$ and
$s(y_n;\widetilde{\mathbf p}_n)\in[0,1]$ imply
\[
0
\le
\frac{s(y_n;\widetilde{\mathbf p}_n)}{r_n}
\le
\frac{1}{\gamma_r}
=
B_S.
\]
This is the only quantitative property of the uncertainty scale used in
the residual-potential argument.

Indeed, the proof of
Lemma~\ref{lem:local_residual_potential} depends on the scaled score only
through its boundedness and the equivalence
\[
y_n\notin\mathcal C_n({\bf z}_n)
\quad\Longleftrightarrow\quad
\frac{s(y_n;\widetilde{\mathbf p}_n)}{r_n}
>
q_n+g_n({\bf z}_n).
\]
Once these properties hold, the subsequent marginal and graph-localized
arguments depend only on the joint online update in
Equation~\eqref{eq:local_joint_update}.

Therefore, Theorems~\ref{thm:marginal} and~\ref{thm:local} hold for any
predictable scale sequence satisfying
$r_n\in[\gamma_r,1]$, regardless of how $r_n$ is constructed.
In particular, the MI-based scale in
Equation~\eqref{eq:mi_scale} is a specific uncertainty-adaptation
mechanism used by GLocalCP, rather than a requirement of the calibration
guarantees.
\end{proof}
\section{Experimental Protocol and Implementation Details}
\label{app:experimental_protocol}

This appendix summarizes the experimental protocol and implementation details
not included in the main text. Unless otherwise stated, all experiments use
$\alpha=0.1$ and 50 seeds. Within each dataset and backbone, all methods share
the same graph partition, predictive checkpoint, query order, and calibration
prefix.

\subsection{Inductive Evaluation Protocol}
\label{app:inductive_protocol}

We evaluate node classification and held-out edge prediction on Cora,
CiteSeer, PubMed, Computers, Roman-empire, Amazon-ratings, and ogbn-arxiv.
For node classification, approximately $10\%$ of the vertices form the
initial training graph and $2\%$ are reserved for validation; for edge
prediction, the corresponding fractions are $20\%$ and $2\%$.

For non-temporal benchmarks, the remaining vertices are randomly ordered
after the initial split. For ogbn-arxiv, vertices arrive chronologically by
publication year, with random tie-breaking within each year. The GNN is
trained only on the initial graph and frozen throughout the stream. When a
new vertex arrives, only support relations available at that time are exposed
to the encoder, so all predictions are inductive and causal.

For edge prediction, true edges are separated into observable support edges
and hidden query edges. Hidden query edges are never exposed to the encoder,
including after their labels are revealed. The main experiments use a
$1{:}10$ positive-to-negative candidate construction; balanced $1{:}1$ and
two-hop hard-negative candidates are used for robustness experiments.
Negatives are checked against the full ground-truth graph to exclude false
negatives.

The first 200 queries of each stream form the common calibration prefix and
are excluded from evaluation. Each subsequent label is revealed only after
the corresponding prediction set has been constructed and may then be used
by adaptive conformal methods.

\subsection{Model and Calibration Settings}
\label{app:implementation_details}

GraphSAGE is the default backbone, with GCN used for the backbone robustness
study. Node models use 64-dimensional hidden representations with dropout
rate $0.5$, while edge models use dropout rate $0.3$. Predictive probabilities
are obtained either deterministically or by averaging MC-dropout predictions,
using 30 stochastic passes for node classification and ten for edge
prediction. Dropout is active in both the encoder and decoder for MC edge
prediction.

For graph-local calibration, node queries use normalized hidden
representations, while edge queries use normalized symmetric representations
of their endpoint embeddings. All representations are constructed before the
corresponding query label is observed.

GLocalCP initializes its global threshold from the common calibration prefix
and its local correction at zero. The default configuration uses MI-based
uncertainty scaling with $\gamma_r=0.4$, step size $\eta=0.05$, and an RBF
kernel with bandwidth $h=0.5$. The scalable controller uses 512 random Fourier
features; the exact-kernel construction and its approximation are described
in \Cref{app:kernel_rff}. The unscaled ablation fixes $r_n=1$.

All predictive models are trained only on the initial graph, selected using
the validation split, and frozen before the query stream begins. We use Adam
with learning rates $0.01$ and $0.005$ for node and edge prediction,
respectively, and weight decay $5\times10^{-4}$ and $10^{-4}$.

\subsection{Baselines and Evaluation}
\label{app:baseline_evaluation}

For node classification, we compare with NAPS-U, DAPS, and SNAPS, restricting
their graph neighborhoods to information available before the current query.
For edge prediction, we compare with CLP-CQR, CLP-SCQR, RR-CQR, and
RR-Mondrian under the same inductive graph construction. Any graph statistics,
residual information, or group assignments required by these methods are
constructed without future query labels or hidden query edges. Native edge
intervals are intersected with $\{0,1\}$ to obtain binary prediction sets,
which preserves the realized binary coverage event.

We report marginal coverage, mean prediction-set size, and graph-local
calibration. For each seed, 64 representations from the calibration prefix
serve as common audit anchors for the graph-local metric defined in
\Cref{sec:local}. All methods use the same anchors and the exact RBF kernel
with bandwidth $h=0.5$, including GLocalCP runs whose online controller uses
random Fourier features. All metrics are computed over the complete post-calibration stream within
each seed and then averaged across 50 seeds. The graph-local maximum is
computed over the audit anchors within each seed before averaging across
seeds.
\section{Kernel Construction and Random-Feature Approximation}
\label{app:kernel_rff}

The graph-local update uses a fixed kernel over normalized graph-query
representations. The theoretical analysis considers the exact kernel, whereas
the experiments use a finite-dimensional random Fourier feature (RFF)
approximation for scalability.

\subsection{Exact Kernel Construction}
\label{app:kernel_constructions}

Let ${\bf z},{\bf z}'\in\mathbb R^d$ denote normalized graph-query
representations. For bandwidth $h>0$, the RBF kernel is
\begin{equation}
    k_h^{\mathrm{RBF}}({\bf z},{\bf z}')
    =
    \exp\left(
        -\frac{\|{\bf z}-{\bf z}'\|_2^2}{2h^2}
    \right).
\label{eq:local_rbf}
\end{equation}
We also consider the Mat\'ern family,
\begin{equation}
    k_{\nu,h}^{\mathrm{Mat}}({\bf z},{\bf z}')
    =
    \frac{2^{1-\nu}}{\Gamma(\nu)}
    \left(
        \frac{\sqrt{2\nu}\|{\bf z}-{\bf z}'\|_2}{h}
    \right)^\nu
    K_\nu\left(
        \frac{\sqrt{2\nu}\|{\bf z}-{\bf z}'\|_2}{h}
    \right),
\label{eq:local_matern}
\end{equation}
with $k_{\nu,h}^{\mathrm{Mat}}({\bf z},{\bf z})=1$ by continuous extension.
Both kernels are fixed before the online stream and are independent of
subsequently revealed labels.

The representations are normalized as
\begin{equation}
    {\bf z}
    =
    \frac{h}{\max\{\|h\|_2,10^{-12}\}},
\label{eq:kernel_covariate_normalization}
\end{equation}
so $\|{\bf z}\|_2\le1$. Hence
$\|{\bf z}-{\bf z}'\|_2\le2$, and the RBF kernel satisfies
\begin{equation}
    k_h^{\mathrm{RBF}}({\bf z},{\bf z}')
    \ge
    \exp\left(-\frac{2}{h^2}\right)>0.
\label{eq:rbf_uniform_lower_bound}
\end{equation}
The Mat\'ern kernel admits an analogous positive lower bound on the same
domain. These properties yield the normalization and uniform positivity used
in Lemma~\ref{lem:kernel_lower_bound}.

All reported experiments use the RBF kernel with bandwidth $h=0.5$. The
bandwidth controls the localization scale, with smaller values concentrating
feedback around more similar representations. The uniform lower bound above
is a worst-case theoretical constant and is not intended to characterize the
empirical graph-local gap.

\subsection{Random Feature Approximation}
\label{app:rff_implementation}

The exact graph-local update requires retaining kernel contributions from
previous queries. To obtain a fixed-dimensional implementation, we approximate
the RBF kernel with random Fourier features.

For even dimension $D$, draw
\[
    \omega_j
    \sim
    \mathcal N(0,h^{-2}I_d),
    \qquad
    j=1,\ldots,D/2,
\]
and define
\begin{equation}
    \phi({\bf z})
    =
    \sqrt{\frac{2}{D}}
    \begin{bmatrix}
        \cos(\omega_1^\top{\bf z}) &
        \cdots &
        \cos(\omega_{D/2}^\top{\bf z}) &
        \sin(\omega_1^\top{\bf z}) &
        \cdots &
        \sin(\omega_{D/2}^\top{\bf z})
    \end{bmatrix}^{\!\top}.
\label{eq:rff_map}
\end{equation}
Then
\[
    \|\phi({\bf z})\|_2^2=1,
    \qquad
    \mathbb E_{\omega}
    \left[
        \phi({\bf z})^\top\phi({\bf z}')
    \right]
    =
    k_h^{\mathrm{RBF}}({\bf z},{\bf z}').
\]

The feature map is sampled before the online stream and remains fixed within
each run. Writing
\[
    g_n({\bf z})
    =
    {\bf w}_n^\top\phi({\bf z}),
\]
the local update becomes
\begin{equation}
    {\bf w}_{n+1}
    =
    {\bf w}_n
    +
    \eta\epsilon_n\phi({\bf z}_n).
\label{eq:rff_weight_update}
\end{equation}
We use $D=512$, $\eta=0.05$, and ${\bf w}_1=0$ in all reported experiments.

\subsection{Relation to the Exact-Kernel Analysis}
\label{app:exact_vs_rff}

Theorem~\ref{thm:local} applies to the exact kernel update under the
normalization and uniform-positivity conditions above. The finite-dimensional
approximation
\[
    \widehat{k}_D({\bf z},{\bf z}')
    =
    \phi({\bf z})^\top\phi({\bf z}')
\]
preserves the normalized diagonal but need not be positive for every pair of
representations. We therefore use RFF as a scalable empirical approximation
and do not claim that the exact-kernel bound automatically transfers to
finite $D$.

This approximation is kept separate from evaluation. As described in
\Cref{app:implementation_details}, the reported graph-local metric $K$ is
computed using the exact RBF kernel with $h=0.5$ at the shared audit anchors,
including for GLocalCP runs whose online controller uses RFF.

Finally, the proof of Theorem~\ref{thm:local} uses zero initialization for a
clean pathwise bound, whereas the experiments initialize the scalar threshold
from the common calibration prefix as described in
\Cref{app:baseline_evaluation}. The local RFF state remains initialized at
zero.
\section{Additional Experimental Results}
\label{app:additional_results}

This appendix provides additional stream-wise calibration results and
robustness evaluations complementing the main experiments.

%==============================================================================
\subsection{Calibration Dynamics Across Datasets}
\label{app:seven_figs}

Figure~\ref{fig:stream_dynamics} in the main text illustrates the evolution
of marginal and graph-local calibration on ogbn-arxiv. Here we report the
same diagnostics on the remaining six benchmarks under the same experimental
protocol.

\begin{figure}[t]
    \centering
    \includegraphics[width=0.7\textwidth]
    {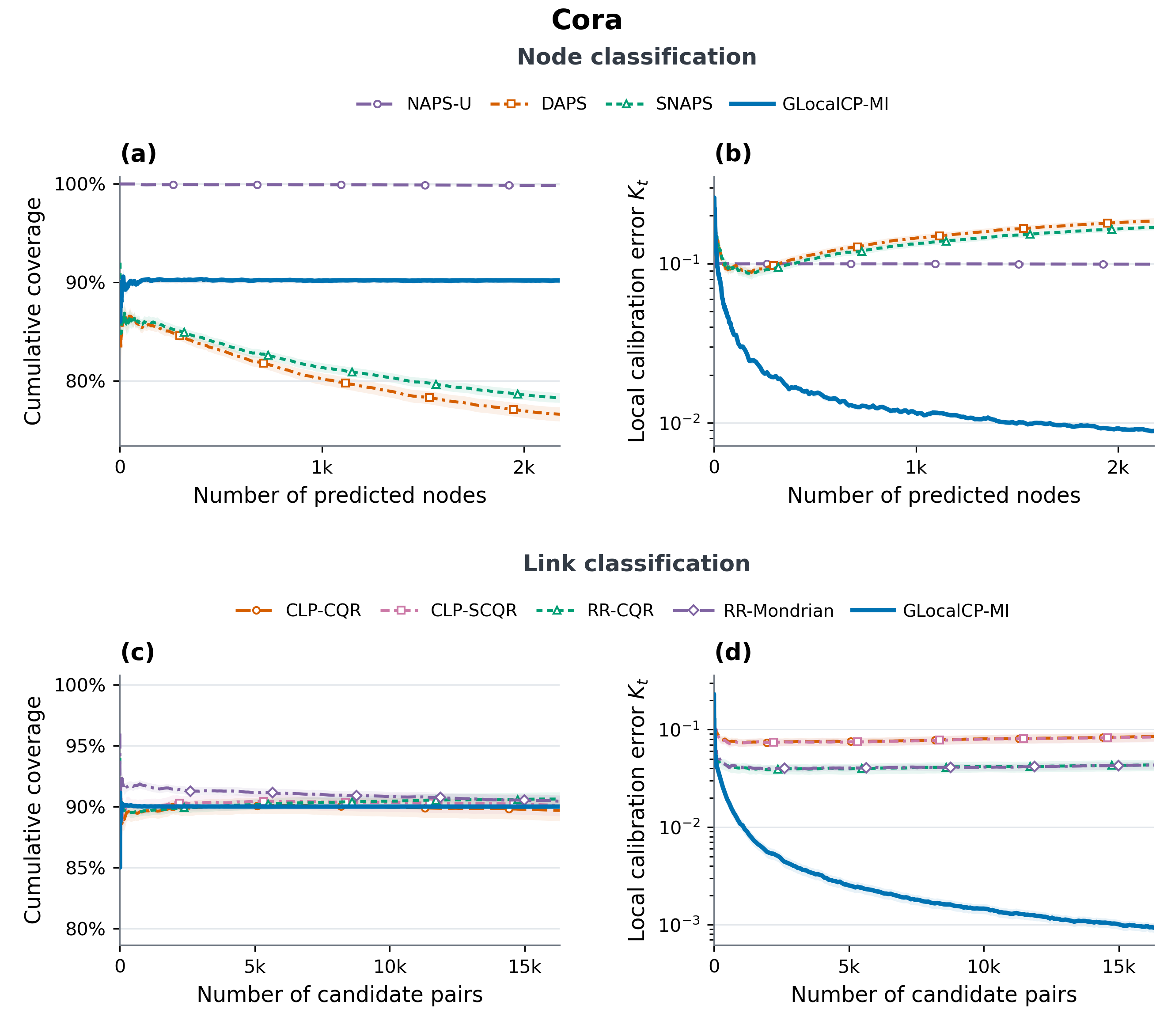}
    \caption{Calibration dynamics on Cora.}
    \label{fig:dynamics-cora}
\end{figure}

\begin{figure}[t]
    \centering
    \includegraphics[width=0.7\textwidth]
    {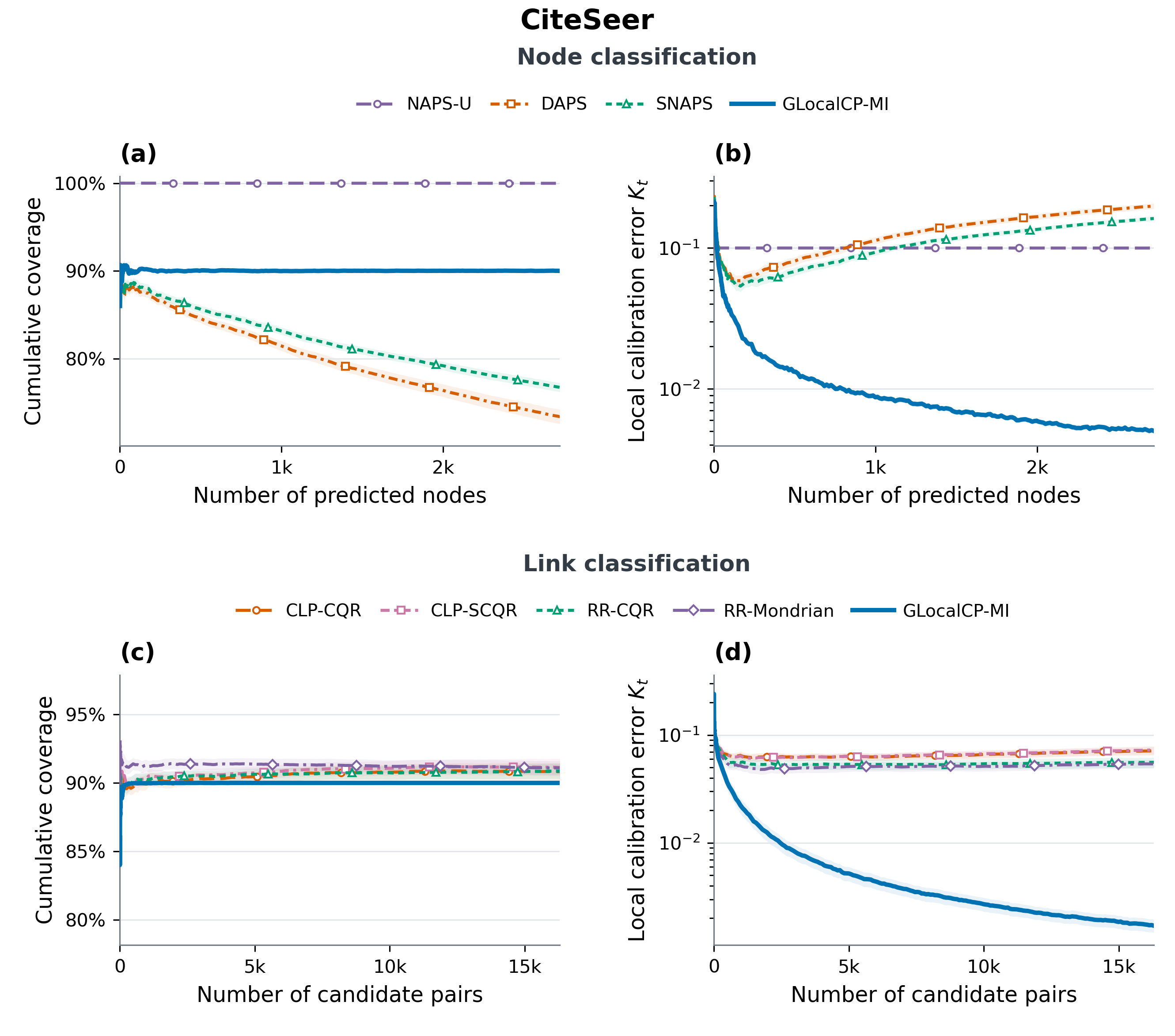}
    \caption{Calibration dynamics on CiteSeer.}
    \label{fig:dynamics-citeseer}
\end{figure}

\begin{figure}[t]
    \centering
    \includegraphics[width=0.7\textwidth]
    {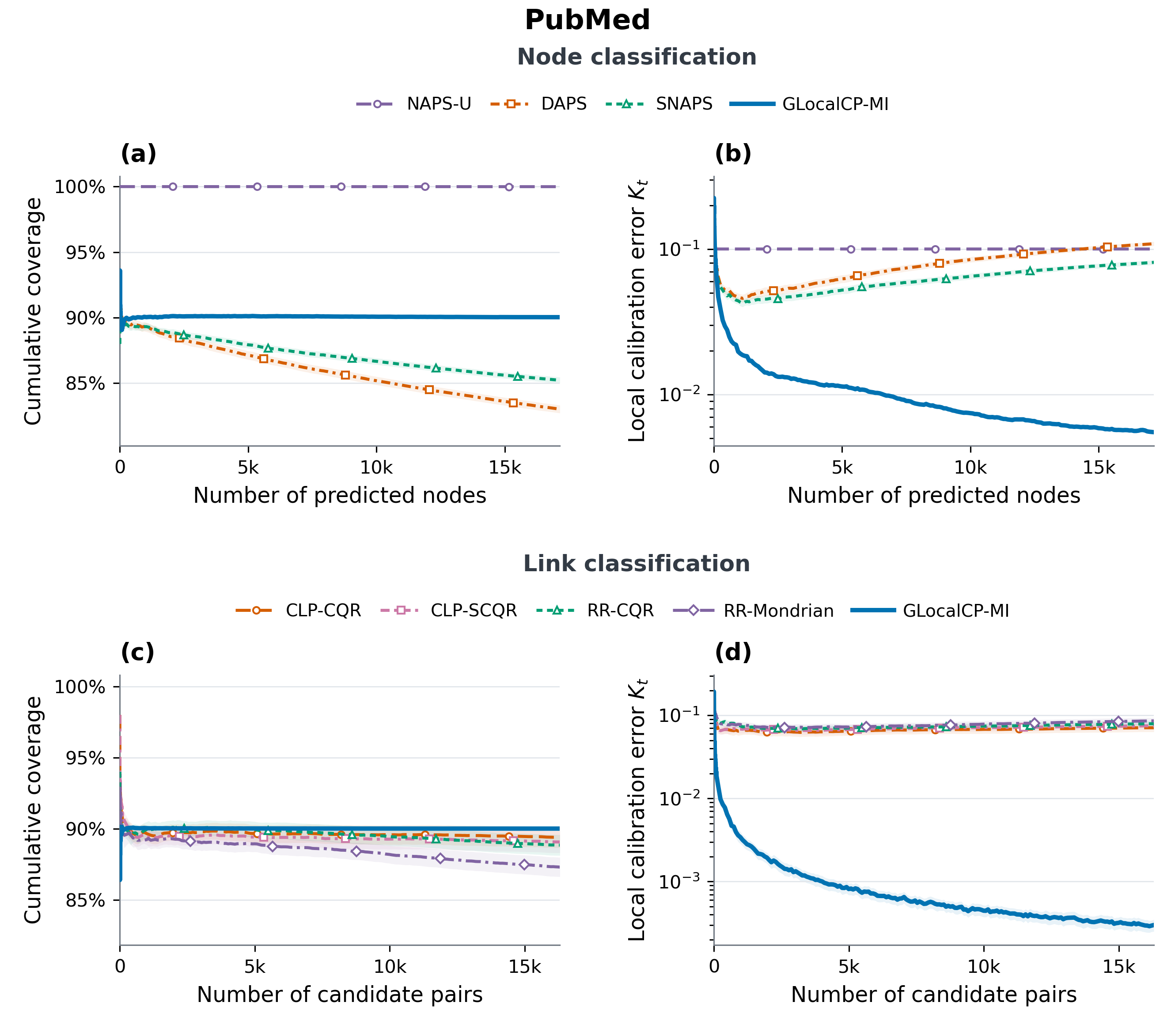}
    \caption{Calibration dynamics on PubMed.}
    \label{fig:dynamics-pubmed}
\end{figure}

\begin{figure}[t]
    \centering
    \includegraphics[width=0.7\textwidth]
    {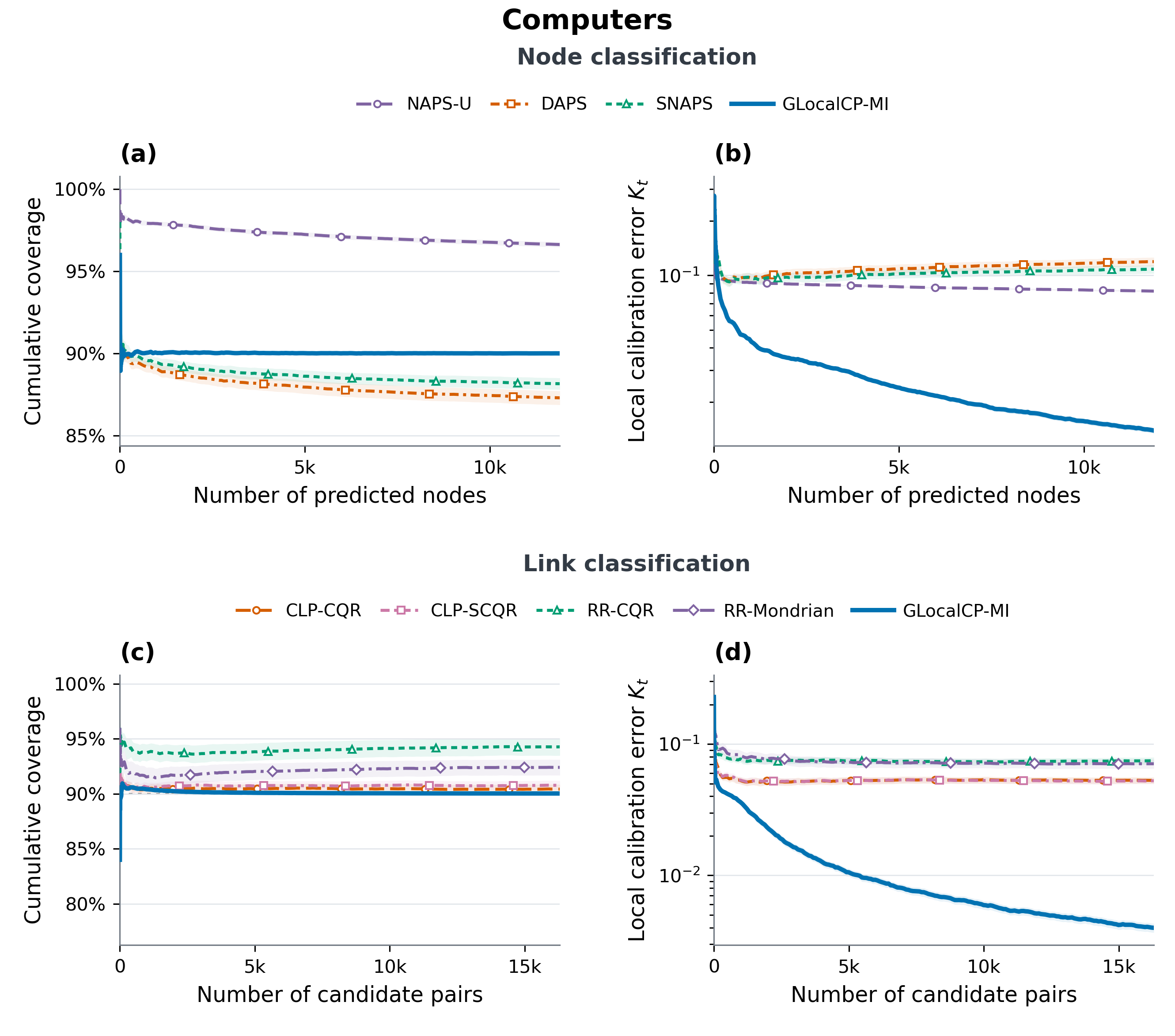}
    \caption{Calibration dynamics on Computers.}
    \label{fig:dynamics-computers}
\end{figure}

\begin{figure}[t]
    \centering
    \includegraphics[width=0.7\textwidth]
    {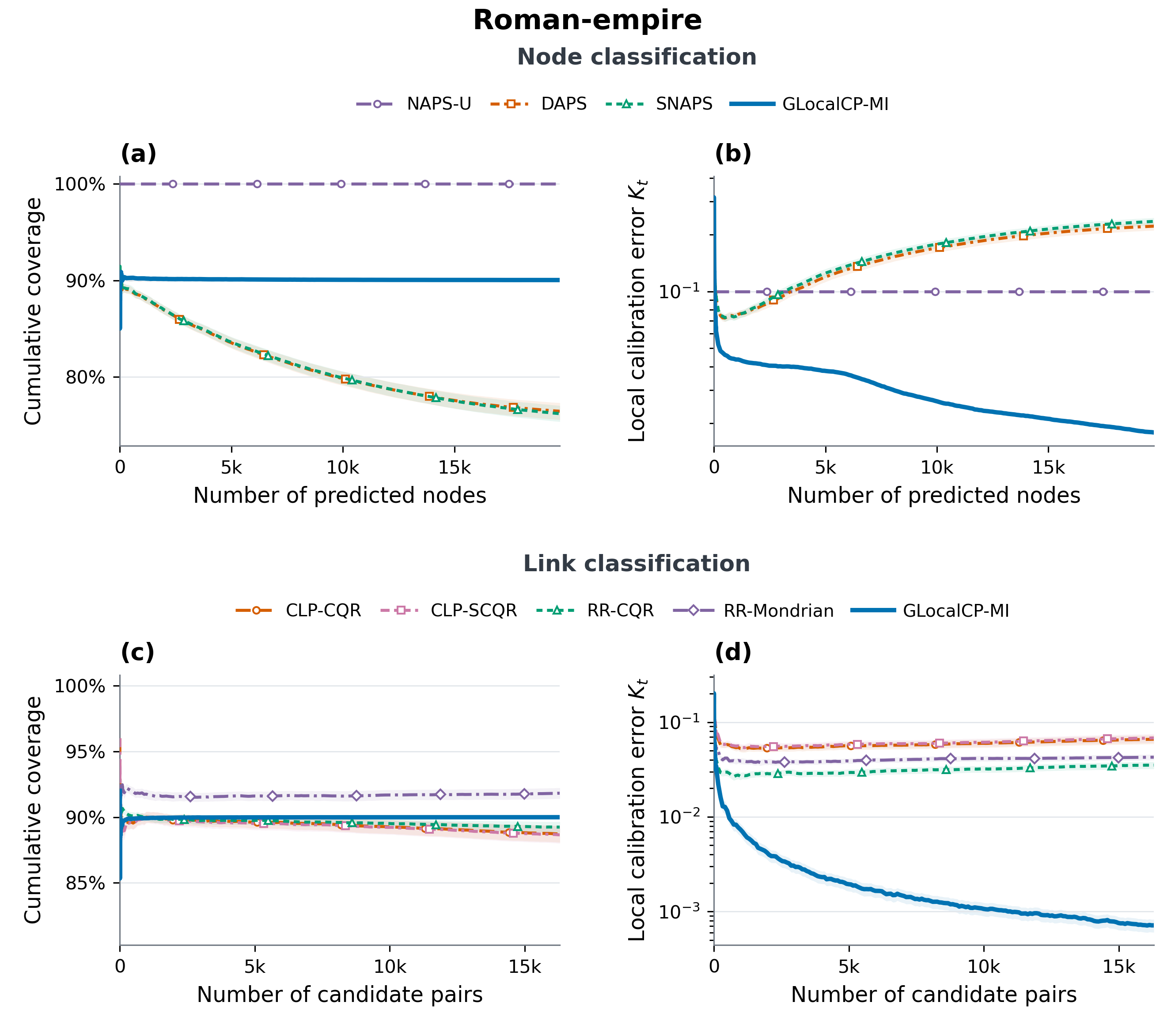}
    \caption{Calibration dynamics on Roman-empire.}
    \label{fig:dynamics-roman}
\end{figure}

\begin{figure}[t]
    \centering
    \includegraphics[width=0.7\textwidth]
    {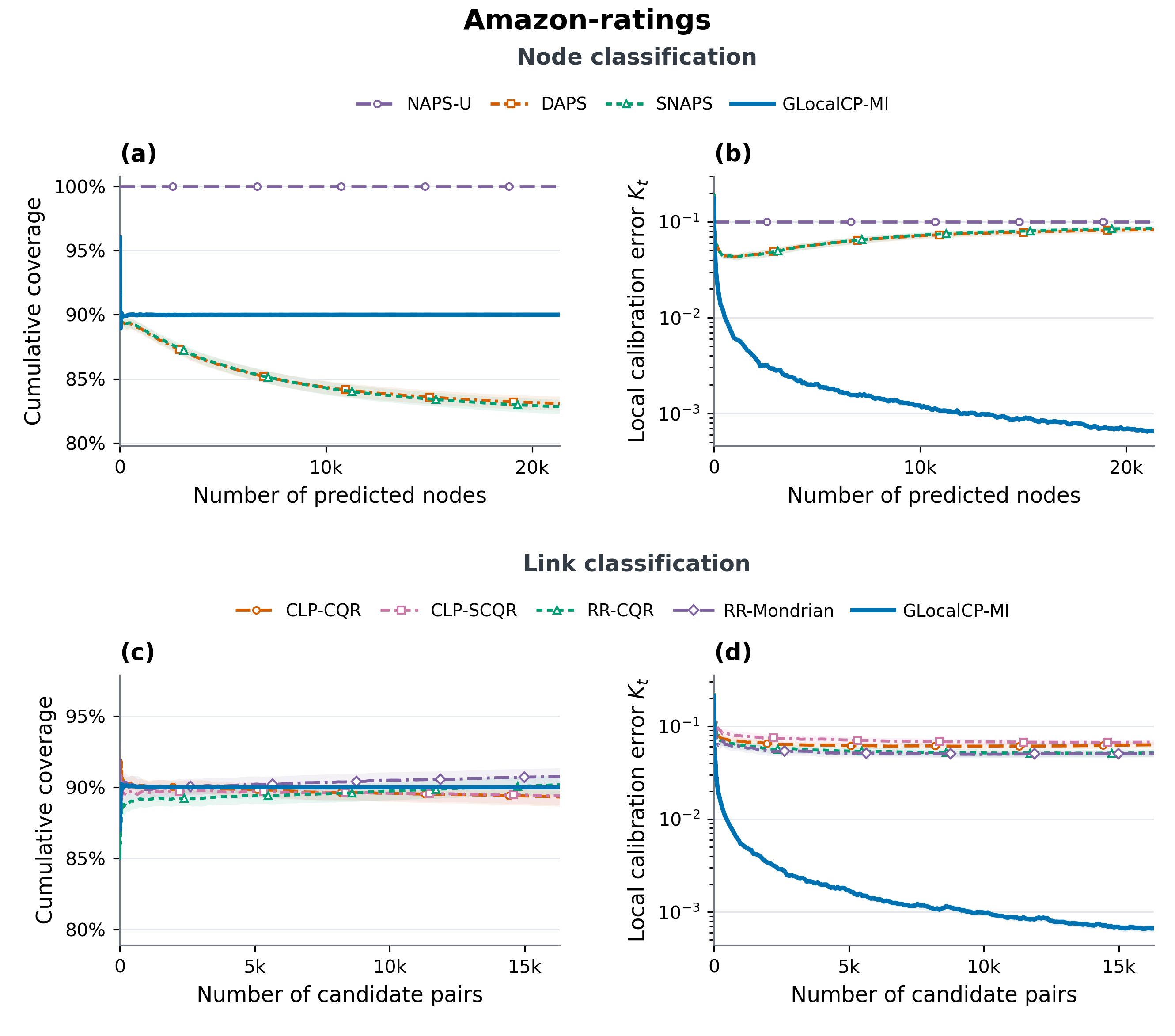}
    \caption{Calibration dynamics on Amazon-ratings.}
    \label{fig:dynamics-amazon}
\end{figure}

Across the six additional benchmarks, the pattern observed on ogbn-arxiv
persists: similar marginal coverage can coexist with substantially different
graph-local calibration. GLocalCP consistently reduces these local
discrepancies while maintaining coverage near the nominal target, indicating
that the marginal--local separation is not specific to a particular graph or
arrival sequence.

%==============================================================================
\subsection{Robustness to Backbone and Conformality Score}
\label{app:updated_experiment_results}

The main experiments use GraphSAGE with LAC. We additionally evaluate
GCN--LAC, GraphSAGE--APS, and GCN--APS under the same experimental protocol.
Tables~\ref{tab:gcn-aps-node} and~\ref{tab:gcn-aps-edge} report the complete
node and edge results, respectively, at target coverage $0.9$ over 50 seeds.
For edge prediction, CLP and RR retain their native CQR-based constructions;
the LAC/APS choice therefore applies only to GLocalCP, and their results are
repeated within each backbone for reference.

\begin{table}[!ht]
\centering
\caption{
Node classification under alternative backbone and conformality-score
configurations. Entries are means over 50 seeds at target coverage $0.9$.
}
\label{tab:gcn-aps-node}
\fontsize{8}{10}\selectfont
\setlength{\tabcolsep}{2.5pt}
\renewcommand{\arraystretch}{1.12}

\begin{tabular}{llrrrrrrrrr}
\toprule
& & \multicolumn{3}{c}{GCN + LAC}
& \multicolumn{3}{c}{GraphSAGE + APS}
& \multicolumn{3}{c}{GCN + APS}\\
\cmidrule(lr){3-5}
\cmidrule(lr){6-8}
\cmidrule(lr){9-11}
Dataset & Method
& Cov. & Size & $K$
& Cov. & Size & $K$
& Cov. & Size & $K$\\
\midrule

Cora
& NAPS-U
& 0.999 & 6.802 & 0.0991
& 0.997 & 6.892 & 0.0988
& 0.996 & 6.901 & 0.0976\\
& DAPS
& 0.792 & 1.934 & 0.1641
& 0.910 & 4.259 & 0.0434
& 0.893 & 4.117 & 0.0486\\
& SNAPS
& 0.810 & 1.797 & 0.1454
& 0.939 & 4.086 & 0.0560
& 0.933 & 3.939 & 0.0520\\
& GLocalCP-MI
& 0.903 & 2.077 & 0.0106
& 0.899 & 4.236 & 0.0074
& 0.900 & 4.555 & 0.0051\\

\addlinespace[3pt]
CiteSeer
& NAPS-U
& 1.000 & 5.977 & 0.0999
& 1.000 & 5.989 & 0.0999
& 0.999 & 5.982 & 0.0994\\
& DAPS
& 0.750 & 2.233 & 0.1817
& 0.870 & 3.775 & 0.0703
& 0.860 & 3.664 & 0.0615\\
& SNAPS
& 0.774 & 2.153 & 0.1565
& 0.902 & 3.760 & 0.0573
& 0.889 & 3.571 & 0.0428\\
& GLocalCP-MI
& 0.900 & 3.021 & 0.0053
& 0.899 & 4.041 & 0.0050
& 0.900 & 4.205 & 0.0035\\

\addlinespace[3pt]
PubMed
& NAPS-U
& 1.000 & 2.999 & 0.1000
& 1.000 & 2.999 & 0.1000
& 1.000 & 2.999 & 0.0999\\
& DAPS
& 0.831 & 1.073 & 0.1172
& 0.790 & 1.634 & 0.1620
& 0.762 & 1.561 & 0.2065\\
& SNAPS
& 0.848 & 1.068 & 0.0955
& 0.776 & 1.593 & 0.1877
& 0.763 & 1.549 & 0.2154\\
& GLocalCP-MI
& 0.900 & 1.202 & 0.0052
& 0.900 & 1.970 & 0.0008
& 0.900 & 1.951 & 0.0010\\

\addlinespace[3pt]
Computers
& NAPS-U
& 0.960 & 5.989 & 0.0775
& 0.960 & 6.563 & 0.0829
& 0.850 & 6.687 & 0.1774\\
& DAPS
& 0.881 & 1.499 & 0.0890
& 0.877 & 2.942 & 0.1595
& 0.706 & 2.412 & 0.3797\\
& SNAPS
& 0.885 & 1.408 & 0.0854
& 0.842 & 2.727 & 0.2388
& 0.691 & 2.166 & 0.4205\\
& GLocalCP-MI
& 0.900 & 1.386 & 0.0123
& 0.900 & 3.787 & 0.0064
& 0.900 & 6.898 & 0.0021\\

\addlinespace[3pt]
Roman-empire
& NAPS-U
& 1.000 & 18.000 & 0.1000
& 1.000 & 18.000 & 0.1000
& 1.000 & 18.000 & 0.1000\\
& DAPS
& 0.756 & 10.272 & 0.2019
& 0.755 & 8.087 & 0.3533
& 0.766 & 10.698 & 0.1985\\
& SNAPS
& 0.711 & 6.246 & 0.2390
& 0.752 & 7.135 & 0.3797
& 0.700 & 7.803 & 0.2701\\
& GLocalCP-MI
& 0.900 & 12.632 & 0.0009
& 0.900 & 6.167 & 0.0042
& 0.900 & 12.363 & 0.0013\\

\addlinespace[3pt]
Amazon-ratings
& NAPS-U
& 1.000 & 5.000 & 0.1000
& 1.000 & 5.000 & 0.1000
& 1.000 & 5.000 & 0.1000\\
& DAPS
& 0.814 & 2.739 & 0.0961
& 0.839 & 2.879 & 0.0664
& 0.800 & 2.690 & 0.1048\\
& SNAPS
& 0.813 & 2.743 & 0.0968
& 0.829 & 2.831 & 0.0780
& 0.792 & 2.657 & 0.1128\\
& GLocalCP-MI
& 0.900 & 3.354 & 0.0005
& 0.900 & 3.390 & 0.0006
& 0.900 & 3.471 & 0.0004\\

\addlinespace[3pt]
ogbn-arxiv
& NAPS-U
& 1.000 & 40.000 & 0.1000
& 1.000 & 40.000 & 0.1000
& 1.000 & 40.000 & 0.1000\\
& DAPS
& 0.847 & 6.078 & 0.0752
& 0.937 & 13.433 & 0.0668
& 0.890 & 10.464 & 0.1044\\
& SNAPS
& 0.848 & 5.987 & 0.0754
& 0.932 & 11.859 & 0.0711
& 0.886 & 9.304 & 0.1145\\
& GLocalCP-MI
& 0.900 & 11.798 & 0.0023
& 0.900 & 8.745 & 0.0015
& 0.900 & 13.062 & 0.0008\\

\bottomrule
\end{tabular}
\end{table}

Across all three configurations, GLocalCP maintains near-target coverage and
substantially smaller graph-local gaps across the seven datasets. Backbone
and score choice can affect prediction-set size, but the graph-local
calibration improvement persists.

\begin{table}[!ht]
\centering
\caption{
Held-out edge prediction with $1{:}10$ candidates under alternative backbone
and conformality-score configurations. Entries are means over 50 seeds at
target coverage $0.9$.
}
\label{tab:gcn-aps-edge}
\fontsize{8}{10}\selectfont
\setlength{\tabcolsep}{2.5pt}
\renewcommand{\arraystretch}{1.12}

\begin{tabular}{llrrrrrrrrr}
\toprule
& & \multicolumn{3}{c}{GCN + LAC}
& \multicolumn{3}{c}{GraphSAGE + APS}
& \multicolumn{3}{c}{GCN + APS}\\
\cmidrule(lr){3-5}
\cmidrule(lr){6-8}
\cmidrule(lr){9-11}
Dataset & Method
& Cov. & Size & $K$
& Cov. & Size & $K$
& Cov. & Size & $K$\\
\midrule

Cora
& CLP-CQR
& 0.879 & 1.793 & 0.0815
& 0.897 & 1.815 & 0.0852
& 0.879 & 1.793 & 0.0815\\
& CLP-SCQR
& 0.880 & 1.791 & 0.0823
& 0.901 & 1.818 & 0.0839
& 0.880 & 1.791 & 0.0823\\
& RR-CQR
& 0.890 & 1.520 & 0.0441
& 0.906 & 1.361 & 0.0433
& 0.890 & 1.520 & 0.0441\\
& RR-Mondrian
& 0.921 & 1.703 & 0.0365
& 0.905 & 1.508 & 0.0432
& 0.921 & 1.703 & 0.0365\\
& GLocalCP-MI
& 0.900 & 1.323 & 0.0031
& 0.900 & 1.455 & 0.0017
& 0.900 & 1.322 & 0.0017\\

\addlinespace[3pt]
CiteSeer
& CLP-CQR
& 0.892 & 1.789 & 0.0791
& 0.908 & 1.784 & 0.0714
& 0.892 & 1.789 & 0.0791\\
& CLP-SCQR
& 0.892 & 1.786 & 0.0775
& 0.911 & 1.784 & 0.0728
& 0.892 & 1.786 & 0.0775\\
& RR-CQR
& 0.904 & 1.674 & 0.0396
& 0.909 & 1.427 & 0.0558
& 0.904 & 1.674 & 0.0396\\
& RR-Mondrian
& 0.912 & 1.723 & 0.0326
& 0.911 & 1.520 & 0.0540
& 0.912 & 1.723 & 0.0326\\
& GLocalCP-MI
& 0.900 & 1.316 & 0.0034
& 0.900 & 1.492 & 0.0015
& 0.900 & 1.320 & 0.0021\\

\addlinespace[3pt]
PubMed
& CLP-CQR
& 0.844 & 1.416 & 0.0988
& 0.894 & 1.620 & 0.0710
& 0.844 & 1.416 & 0.0988\\
& CLP-SCQR
& 0.843 & 1.432 & 0.1024
& 0.891 & 1.595 & 0.0748
& 0.843 & 1.432 & 0.1024\\
& RR-CQR
& 0.804 & 1.696 & 0.1151
& 0.888 & 1.720 & 0.0793
& 0.804 & 1.696 & 0.1151\\
& RR-Mondrian
& 0.823 & 1.726 & 0.0941
& 0.873 & 1.723 & 0.0859
& 0.823 & 1.726 & 0.0941\\
& GLocalCP-MI
& 0.900 & 1.283 & 0.0038
& 0.900 & 1.339 & 0.0014
& 0.900 & 1.389 & 0.0007\\

\addlinespace[3pt]
Computers
& CLP-CQR
& 0.823 & 1.124 & 0.1222
& 0.904 & 1.755 & 0.0528
& 0.823 & 1.124 & 0.1222\\
& CLP-SCQR
& 0.826 & 1.097 & 0.1204
& 0.908 & 1.715 & 0.0525
& 0.826 & 1.097 & 0.1204\\
& RR-CQR
& 0.924 & 1.808 & 0.1111
& 0.943 & 1.805 & 0.0745
& 0.924 & 1.808 & 0.1111\\
& RR-Mondrian
& 0.835 & 1.749 & 0.1167
& 0.924 & 1.860 & 0.0707
& 0.835 & 1.749 & 0.1167\\
& GLocalCP-MI
& 0.900 & 1.261 & 0.0054
& 0.900 & 1.413 & 0.0015
& 0.900 & 1.681 & 0.0005\\

\addlinespace[3pt]
Roman-empire
& CLP-CQR
& 0.904 & 1.820 & 0.0604
& 0.887 & 1.799 & 0.0660
& 0.904 & 1.820 & 0.0604\\
& CLP-SCQR
& 0.901 & 1.817 & 0.0611
& 0.887 & 1.799 & 0.0681
& 0.901 & 1.817 & 0.0611\\
& RR-CQR
& 0.891 & 1.787 & 0.0464
& 0.892 & 1.404 & 0.0352
& 0.891 & 1.787 & 0.0464\\
& RR-Mondrian
& 0.895 & 1.808 & 0.0455
& 0.918 & 1.602 & 0.0427
& 0.895 & 1.808 & 0.0455\\
& GLocalCP-MI
& 0.900 & 1.456 & 0.0010
& 0.900 & 1.386 & 0.0010
& 0.900 & 1.323 & 0.0021\\

\addlinespace[3pt]
Amazon-ratings
& CLP-CQR
& 0.888 & 1.756 & 0.0761
& 0.893 & 1.846 & 0.0629
& 0.888 & 1.756 & 0.0761\\
& CLP-SCQR
& 0.885 & 1.737 & 0.0755
& 0.894 & 1.831 & 0.0672
& 0.885 & 1.737 & 0.0755\\
& RR-CQR
& 0.851 & 1.770 & 0.0771
& 0.902 & 1.635 & 0.0513
& 0.851 & 1.770 & 0.0771\\
& RR-Mondrian
& 0.862 & 1.793 & 0.0608
& 0.908 & 1.688 & 0.0507
& 0.862 & 1.793 & 0.0608\\
& GLocalCP-MI
& 0.900 & 1.132 & 0.0032
& 0.900 & 1.353 & 0.0018
& 0.900 & 1.123 & 0.0026\\

\addlinespace[3pt]
ogbn-arxiv
& CLP-CQR
& 0.775 & 1.202 & 0.1521
& 0.882 & 1.774 & 0.1001
& 0.775 & 1.202 & 0.1521\\
& CLP-SCQR
& 0.790 & 1.155 & 0.1389
& 0.883 & 1.713 & 0.0977
& 0.790 & 1.155 & 0.1389\\
& RR-CQR
& 0.820 & 1.738 & 0.1001
& 0.940 & 1.852 & 0.0906
& 0.820 & 1.738 & 0.1001\\
& RR-Mondrian
& 0.851 & 1.786 & 0.0732
& 0.920 & 1.842 & 0.0664
& 0.851 & 1.786 & 0.0732\\
& GLocalCP-MI
& 0.899 & 1.219 & 0.0036
& 0.900 & 1.300 & 0.0026
& 0.900 & 1.656 & 0.0003\\

\bottomrule
\end{tabular}

\par\medskip
\begin{minipage}{\linewidth}
\footnotesize
Size denotes binary prediction-set cardinality after intersection with
$\{0,1\}$.
\end{minipage}
\end{table}

The edge results show the same pattern: GLocalCP maintains coverage near the
target and graph-local gaps on the order of $10^{-3}$ across the alternative
configurations, while the native edge baselines retain substantially larger
local discrepancies. Together with the node results, these experiments show
that the graph-local calibration improvement is robust to both backbone and
conformality-score choice.

%==============================================================================

\section{Additional Ablations and Robustness}
\label{app:additional_ablations}

This appendix isolates the mechanisms responsible for the empirical behavior
of GLocalCP and examines whether its conclusions persist under changes to the
query distribution and predictive uncertainty estimator. We focus on three
questions: which component drives the improvement in graph-local calibration,
how the method behaves under alternative edge-candidate distributions, and
whether the observed behavior depends on MC prediction. Unless stated
otherwise, all experiments use GraphSAGE, MC prediction, LAC,
$\alpha=0.1$, and 50 seeds.

\subsection{Local Feedback and Uncertainty Scaling}
\label{app:component_ablation}

We isolate the two design choices in GLocalCP through controlled comparisons
that vary one component at a time. We first examine the kernel-local feedback,
which is the mechanism introduced to address calibration disparities that are
not controlled by marginal coverage. We compare GLocalCP-MI with Global MI:
both use the same MI-dependent scale, but Global MI removes the local
correction by setting $g_n\equiv0$. We then examine uncertainty scaling by
comparing GLocalCP-MI with Local $r=1$: both retain the same kernel-local
feedback, but the latter fixes $r_n=1$. The predictive probabilities, graph
stream, calibration prefix, and remaining online updates are held fixed in
each comparison.

The two comparisons therefore test complementary aspects of the method.
Global MI versus GLocalCP-MI isolates whether $g_n$ reduces graph-local
calibration error beyond a globally adapted controller with the same
uncertainty scale. Local $r=1$ versus GLocalCP-MI instead tests whether
uncertainty-dependent scaling improves prediction-set efficiency while
preserving marginal coverage.

\paragraph{Node classification.}
Table~\ref{tab:ablation-node-controlled} reports the two controlled comparisons
for node classification. The left block isolates local feedback through the
graph-local gap $K$. The right block isolates uncertainty scaling through
marginal coverage and prediction-set size.

\begin{table*}[t]
\centering
\scriptsize
\setlength{\tabcolsep}{4pt}
\renewcommand{\arraystretch}{1.12}
\caption{
Controlled component ablations for node classification with GraphSAGE, MC
prediction, and LAC at $\alpha=0.1$, averaged over 50 seeds. The left block
isolates local feedback by comparing Global MI with GLocalCP-MI. The right
block isolates uncertainty scaling by comparing Local $r=1$ with
GLocalCP-MI. Coverage is reported to three decimal places.
}
\label{tab:ablation-node-controlled}
\begin{tabular}{lcccccc}
\toprule
& \multicolumn{2}{c}{Effect of local feedback}
& \multicolumn{4}{c}{Effect of uncertainty scaling} \\
\cmidrule(lr){2-3}\cmidrule(lr){4-7}
Dataset
& Global MI $K$
& GLocalCP-MI $K$
& Local $r=1$ Cov.
& GLocalCP-MI Cov.
& Local $r=1$ Size
& GLocalCP-MI Size \\
\midrule
Cora
& 0.0159 & 0.0089
& 0.902 & 0.902 & 2.631 & 2.527 \\
CiteSeer
& 0.0121 & 0.0050
& 0.900 & 0.900 & 3.081 & 2.972 \\
PubMed
& 0.0209 & 0.0055
& 0.900 & 0.900 & 1.189 & 1.156 \\
Computers
& 0.0621 & 0.0139
& 0.900 & 0.900 & 1.597 & 1.473 \\
Roman-empire
& 0.0471 & 0.0179
& 0.900 & 0.900 & 6.242 & 5.363 \\
Amazon-ratings
& 0.0062 & 0.0007
& 0.900 & 0.900 & 3.480 & 3.347 \\
ogbn-arxiv
& 0.0145 & 0.0028
& 0.900 & 0.900 & 13.879 & 10.563 \\
\bottomrule
\end{tabular}
\end{table*}

With the MI scale held fixed, introducing kernel-local feedback reduces $K$
on every node dataset. The reduction is substantial on datasets with larger
global local-calibration gaps: $K$ decreases from $0.0621$ to $0.0139$ on
Computers, from $0.0471$ to $0.0179$ on Roman-empire, and from $0.0145$ to
$0.0028$ on ogbn-arxiv. Since Global MI and GLocalCP-MI use the same
uncertainty scale, this comparison directly attributes the reduction in
graph-local disparity to the local correction $g_n$ rather than to
uncertainty scaling.

Once local feedback is held fixed, MI scaling plays a different role.
GLocalCP-MI preserves the marginal coverage attained by Local $r=1$ while
reducing average prediction-set size on all seven datasets. The reduction is
modest on PubMed, from $1.189$ to $1.156$, but becomes more pronounced on
larger multiclass problems: the mean size decreases from $6.242$ to $5.363$
on Roman-empire and from $13.879$ to $10.563$ on ogbn-arxiv. These results
support the use of MI scaling as an efficiency mechanism rather than as the
source of the local-calibration improvement.

\paragraph{Held-out edge prediction.}
Table~\ref{tab:ablation-edge-controlled} repeats the same controlled
comparisons for held-out edge prediction under the $1{:}10$ candidate
distribution.

\begin{table*}[t]
\centering
\scriptsize
\setlength{\tabcolsep}{4pt}
\renewcommand{\arraystretch}{1.12}
\caption{
Controlled component ablations for held-out edge prediction under the
$1{:}10$ candidate distribution with GraphSAGE, MC prediction, and LAC at
$\alpha=0.1$, averaged over 50 seeds. The left block isolates local feedback,
and the right block isolates uncertainty scaling. Coverage is reported to
three decimal places.
}
\label{tab:ablation-edge-controlled}
\begin{tabular}{lcccccc}
\toprule
& \multicolumn{2}{c}{Effect of local feedback}
& \multicolumn{4}{c}{Effect of uncertainty scaling} \\
\cmidrule(lr){2-3}\cmidrule(lr){4-7}
Dataset
& Global MI $K$
& GLocalCP-MI $K$
& Local $r=1$ Cov.
& GLocalCP-MI Cov.
& Local $r=1$ Size
& GLocalCP-MI Size \\
\midrule
Cora
& 0.0386 & 0.0009
& 0.900 & 0.900 & 1.643 & 1.559 \\
CiteSeer
& 0.0573 & 0.0017
& 0.900 & 0.900 & 1.650 & 1.581 \\
PubMed
& 0.0210 & 0.0003
& 0.900 & 0.900 & 1.551 & 1.454 \\
Computers
& 0.0453 & 0.0040
& 0.900 & 0.900 & 1.396 & 1.372 \\
Roman-empire
& 0.0261 & 0.0007
& 0.900 & 0.900 & 1.641 & 1.542 \\
Amazon-ratings
& 0.0147 & 0.0007
& 0.900 & 0.900 & 1.536 & 1.437 \\
ogbn-arxiv
& 0.0247 & 0.0021
& 0.900 & 0.900 & 1.389 & 1.331 \\
\bottomrule
\end{tabular}
\end{table*}

The edge results make the contribution of local feedback especially clear.
With the same MI scale, Global MI retains graph-local gaps on the order of
$10^{-2}$, whereas GLocalCP-MI reduces $K$ to the $10^{-3}$ scale or below
across all seven datasets. For example, $K$ decreases from $0.0386$ to
$0.0009$ on Cora, from $0.0573$ to $0.0017$ on CiteSeer, and from $0.0210$
to $0.0003$ on PubMed. The reduction therefore persists across all seven
held-out edge streams and cannot be explained by a change in uncertainty
scaling.

The uncertainty-scaling comparison again exhibits a different pattern.
Both Local $r=1$ and GLocalCP-MI attain marginal coverage that rounds to
$0.900$ on every edge dataset, while MI scaling reduces mean prediction-set
size in all seven cases. The reduction ranges from $1.396$ to $1.372$ on
Computers to approximately $0.10$ labels on PubMed, Roman-empire, and
Amazon-ratings. Thus the role of MI scaling is primarily to improve the
efficiency of the locally calibrated prediction sets while retaining their
marginal coverage.

These controlled comparisons mirror the construction of GLocalCP. Starting
from a globally adapted controller with the same uncertainty scale, adding
$g_n$ substantially reduces graph-local calibration error. Once this local
feedback is in place, MI scaling reduces prediction-set size while preserving
marginal coverage. The ablation therefore separates the mechanism that
addresses graph-local disparities from the mechanism that allocates
prediction-set width according to predictive uncertainty.

\subsection{Robustness to Edge Candidate Distributions}
\label{app:robustness}

We next test whether the graph-local improvement persists when the edge-query
distribution changes. In addition to the standard $1{:}10$ construction, we
consider balanced $1{:}1$ candidates and hard negatives sampled preferentially
from two-hop nonedges, as described in \Cref{app:baseline_evaluation}. Each candidate
distribution uses its own calibration prefix, while the graph partition,
predictive checkpoint, and online protocol remain fixed within a seed.

Because the native edge baselines use different prediction-set constructions,
we summarize robustness through the graph-local gap rather than through set
size alone. For each dataset and candidate distribution, we compare
GLocalCP-MI with the smallest mean $K$ among CLP-CQR, CLP-SCQR, RR-CQR, and
RR-Mondrian. We report the relative reduction
\begin{equation}
    \Delta_K
    =
    100\left(
        1-
        \frac{K_{\mathrm{GLocalCP}}}
             {\min_{b\in\mathcal B}K_b}
    \right),
\label{eq:relative-k-reduction}
\end{equation}
where $\mathcal B$ denotes the four native edge baselines. The denominator is
selected independently within each dataset and candidate setting and is used
only as a descriptive reference. The reported quantities are ratios of
50-seed mean gaps, rather than means of per-seed ratios.

\begin{figure}[t]
    \centering
    \includegraphics[width=\linewidth]
    {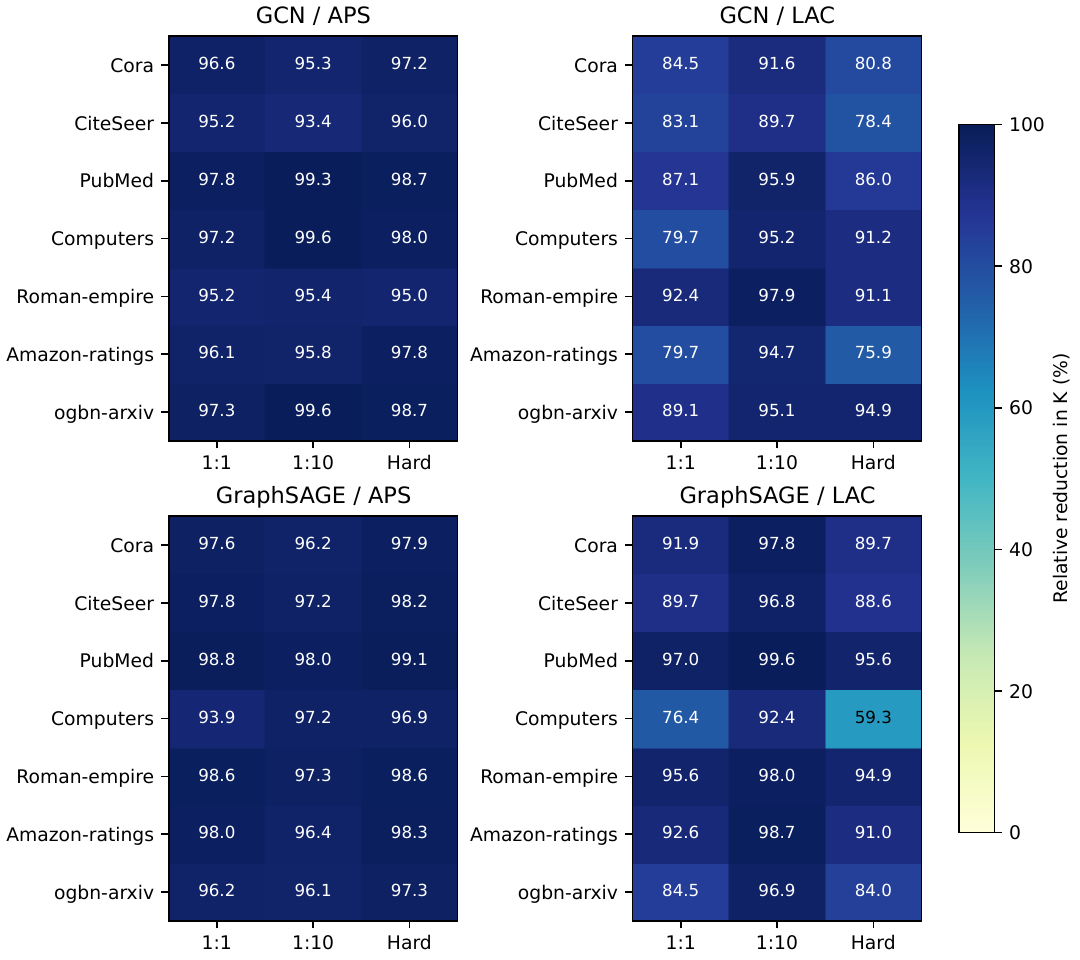}
    \caption{
    Robustness of graph-local calibration to the edge-candidate distribution.
    Each cell reports the relative reduction $\Delta_K$ of GLocalCP-MI against
    the smallest mean graph-local gap among the native edge baselines.
    Results cover both GCN and GraphSAGE backbones, APS and LAC scores, and
    balanced $1{:}1$, standard $1{:}10$, and hard two-hop candidate
    distributions, using all 50 seeds.
    }
    \label{fig:robustness-current}
\end{figure}

The reduction remains positive across all reported datasets, backbones,
scores, and candidate distributions, showing that the local-calibration
advantage is not tied to the standard $1{:}10$ construction. The effect is
particularly stable with APS, for which the reduction remains large across
all three candidate distributions. LAC exhibits greater sensitivity to the
query distribution, especially under hard negatives, but the graph-local gap
of GLocalCP remains below that of the strongest native baseline in every
reported configuration.

To make this distributional sensitivity explicit, Table~\ref{tab:robustness-lac}
reports the GraphSAGE--LAC slice of the heatmap. This is the configuration
closest to the primary experiments in the main text and therefore isolates the
effect of changing only the edge-candidate distribution.

\begin{table}[t]
\centering
\small
\setlength{\tabcolsep}{5pt}
\renewcommand{\arraystretch}{1.12}
\caption{
Relative reduction in graph-local gap $K$ for GraphSAGE with LAC across edge
candidate distributions. Values are percentages computed from 50-seed mean
gaps according to \eqref{eq:relative-k-reduction}.
}
\label{tab:robustness-lac}
\begin{tabular}{lrrr}
\toprule
Dataset & $1{:}1$ & $1{:}10$ & Hard \\
\midrule
Cora           & 91.90 & 97.84 & 89.71 \\
CiteSeer       & 89.72 & 96.84 & 88.57 \\
PubMed         & 97.02 & 99.58 & 95.64 \\
Computers      & 76.45 & 92.40 & 59.30 \\
Roman-empire   & 95.64 & 97.97 & 94.93 \\
Amazon-ratings & 92.61 & 98.69 & 91.04 \\
ogbn-arxiv     & 84.51 & 96.90 & 83.98 \\
\bottomrule
\end{tabular}
\end{table}

The standard $1{:}10$ distribution yields the largest reduction in most
GraphSAGE--LAC experiments, while balanced and hard candidates make the local
calibration problem more demanding. The degradation is most pronounced on
Computers, where the reduction decreases from $92.40\%$ under $1{:}10$
candidates to $59.30\%$ under hard negatives. Nevertheless, the reduction
remains positive even in this setting. PubMed and Roman-empire are
substantially more stable, retaining reductions above $94\%$ under hard
negatives.

These results distinguish robustness from invariance. Changing the candidate
distribution can alter the magnitude of the local-calibration improvement,
sometimes substantially, but does not eliminate it in the completed
experiments. The hard-negative setting therefore provides a stricter test of
local calibration rather than simply reproducing the behavior observed under
random candidate sampling.

\subsection{Alternative Uncertainty Scales}
\label{app:alternative_uncertainty}

The component ablation in Table~\ref{tab:component_ablation} shows that
uncertainty scaling primarily affects prediction-set efficiency, whereas the
local feedback drives the reduction in graph-local calibration error. We
further examine whether this efficiency effect depends specifically on mutual
information (MI). In addition to the unscaled local controller and the default
MI-scaled variant, we evaluate a more aggressive alternative, denoted
GLocalCP-Max, which combines the available uncertainty signals by taking
their maximum.

Across both node and edge prediction, GLocalCP-Max provides no consistent
improvement over MI scaling. A more aggressive scale can change prediction-set
size, but its effect on graph-local calibration varies across datasets and
does not exhibit a systematic advantage over the simpler MI-based rule. These
differences are also substantially smaller than those induced by removing the
graph-local feedback.

We therefore use MI scaling as the default experimental choice. It provides a
single predictive-uncertainty signal without requiring a more aggressive
combination of heterogeneous uncertainty measures. This choice is not required
for the local calibration mechanism: as shown in
Table~\ref{tab:component_ablation}, the local controller provides the dominant
reduction in $K$, while MI scaling primarily improves prediction-set
efficiency.

\subsection{Single versus MC Prediction}
\label{app:single_vs_mc}

Finally, we compare MC dropout with single-pass deterministic prediction.
Averaged across the seven datasets, both settings achieve essentially the
same marginal coverage, remaining near the $0.9$ target for both node and
edge prediction. MC prediction yields a modest but consistent improvement in
prediction-set efficiency: the average set size decreases from $3.960$ to
$3.914$ for node classification and from $1.499$ to $1.468$ for edge
prediction. Thus, incorporating predictive uncertainty through MC dropout
improves set efficiency on average without materially affecting marginal
coverage.
\end{document}